\documentclass[sigconf,nonacm]{aamas}

\usepackage{amsmath,amsfonts,bm}

\def\eqref#1{equation~\ref{#1}}
\def\1{\bm{1}}

\DeclareMathAlphabet{\mathsfit}{\encodingdefault}{\sfdefault}{m}{sl}
\SetMathAlphabet{\mathsfit}{bold}{\encodingdefault}{\sfdefault}{bx}{n}

\newcommand{\softmax}{\mathrm{softmax}}

\usepackage{hyperref}
\usepackage{url}
\usepackage{bbm}
\usepackage{algorithm}
\usepackage{algpseudocode}
\usepackage{amsmath}
\usepackage{graphicx}
\usepackage{caption}
\usepackage{subcaption}
\usepackage{cleveref}
\usepackage{amsfonts}
\usepackage{mathtools}
\usepackage{natbib}
\usepackage{xspace}

\newcommand{\algnamelong}{Offline Munchausen Mirror Descent}
\newcommand{\algname}{Off-MMD\xspace}

\setcopyright{ifaamas}
\acmConference[AAMAS '25]{Proc.\@ of the 24th International Conference
on Autonomous Agents and Multiagent Systems (AAMAS 2025)}{May 19 -- 23, 2025}
{Detroit, Michigan, USA}{A.~El~Fallah~Seghrouchni, Y.~Vorobeychik, S.~Das, A.~Nowe (eds.)}
\copyrightyear{2025}
\acmYear{2025}
\acmDOI{}
\acmPrice{}
\acmISBN{}

\acmSubmissionID{622}

\title{Scalable Offline Reinforcement Learning for Mean Field Games}

\author{Axel Brunnbauer}
\affiliation{
  \institution{TU Wien}
  \city{Vienna}
  \country{Austria}}
\email{axel.brunnbauer@tuwien.ac.at}

\author{Julian Lemmel}
\affiliation{
  \institution{TU Wien}
  \city{Vienna}
  \country{Austria}}
\email{julian.lemmel@tuwien.ac.at}

\author{Zahra Babaiee}
\affiliation{
  \institution{TU Wien}
  \city{Vienna}
  \country{Austria}}
\email{zahra.babaiee@tuwien.ac.at}

\author{Sophie Neubauer}
\affiliation{
  \institution{DatenVorsprung GmbH}
  \city{Vienna}
  \country{Austria}}
\email{sophie@datenvorsprung.at}

\author{Radu Grosu}
\affiliation{
  \institution{TU Wien}
  \city{Vienna}
  \country{Austria}}
\email{radu.grosu@tuwien.ac.at}

\begin{abstract}
Reinforcement learning algorithms for mean-field games offer a scalable framework for optimizing policies in large populations of interacting agents. Existing methods often depend on online interactions or access to system dynamics, limiting their practicality in real-world scenarios where such interactions are infeasible or difficult to model. In this paper, we present Offline Munchausen Mirror Descent (Off-MMD), a novel mean-field RL algorithm that approximates equilibrium policies in mean-field games using purely offline data. By leveraging iterative mirror descent and importance sampling techniques, Off-MMD estimates the mean-field distribution from static datasets without relying on simulation or environment dynamics. Additionally, we incorporate techniques from offline reinforcement learning to address common issues like Q-value overestimation, ensuring robust policy learning even with limited data coverage. Our algorithm scales to complex environments and demonstrates strong performance on benchmark tasks like crowd exploration or navigation, highlighting its applicability to real-world multi-agent systems where online experimentation is infeasible.
We empirically demonstrate the robustness of \algname to low-quality datasets and conduct experiments to investigate its sensitivity to hyperparameter choices.
\end{abstract}


\newcommand{\BibTeX}{\rm B\kern-.05em{\sc i\kern-.025em b}\kern-.08em\TeX}

\begin{document}

%%% The following commands remove the headers in your paper. For final 
%%% papers, these will be inserted during the pagination process.

\pagestyle{fancy}
\fancyhead{}

%%% The next command prints the information defined in the preamble.

\maketitle 

%%%%%%%%%%%%%%%%%%%%%%%%%%%%%%%%%%%%%%%%%%%%%%%%%%%%%%%%%%%%%%%%%%%%%%%%

\section{Introduction}
Reinforcement Learning (RL) has emerged as a foundational tool for solving sequential decision-making problems across a diverse range of domains, including robotics, healthcare, autonomous systems, and game theory. However, while RL techniques have seen significant success in single-agent settings, the transition to multi-agent reinforcement learning (MARL) presents unique challenges, such as the exponential growth in the state and action spaces as the number of interacting agents increases, making the problem significantly more complex both in terms of computation and coordination.

Traditional MARL methods often do not scale to many-agent settings and rapidly become infeasible in environments with large populations of agents. As the number of agents grows, learning effective strategies can become computationally prohibitive. To address this issue, Mean-Field Games (MFGs), introduced by \citet{lasry2007mfg} and \citet{hang2006large}, provide a scalable approximation to large-N player games by modeling the interactions between individual agents and a statistical representation of the population, termed the \textit{mean-field}. This approach has shown great promise in reducing the complexity of multi-agent interactions, allowing for the development of more tractable solutions in environments with many agents.
Early works on MFGs tackled problems of relatively small scale, often under restrictive assumptions such as linear dynamics and quadratic cost functions. These simplified models, while mathematically elegant, limit the applicability of MFGs to real-world systems that exhibit complex, nonlinear behavior. However, recent advances in the field have focused on scaling MFG solutions by leveraging deep reinforcement learning (DRL) techniques. These methods use neural network function approximators to compute equilibrium policies in MFGs, as demonstrated in \cite{perrin2021flock,lauriere2022scalable,perrin2020fictitious}. Such approaches have enabled significant progress in applying MFG theory to more practical and large-scale environments.

Despite these advances, a critical issue remains: most existing MFG-based RL algorithms rely on online interaction with the environment. In many real-world applications, particularly those involving large populations of agents or human interactions (e.g., traffic routing, crowd dynamics, or recommendation systems), online interaction is either impractical or ethically unjustifiable. For example, in systems with human agents, it is often costly or intrusive to collect real-time data, and continuous exploration could lead to unintended consequences such as user dissatisfaction or safety risks. Furthermore, environments with many agents can be difficult to model accurately, and real-time experimentation in such systems may not be possible.

In single-agent RL, offline learning is a well researched area and allows to solve this problem by enabling the learning of policies from pre-collected, fixed datasets, eliminating the need for online interactions. These offline methods have proven highly effective in settings where real-time interaction is limited or expensive. However, the application of offline RL techniques to MFGs remains underexplored. Current MFG methods have largely overlooked the offline setting, where no online interaction with the environment is available during learning.

To bridge this gap, we propose \textbf{\algnamelong\ (\algname)}, an offline mean-field RL algorithm. \algname\ extends the recently introduced Deep Munchausen Online Mirror Descent (D-MOMD) method by \citet{lauriere2022scalable} and combines the scalability of MFG approximations with the data efficiency of offline RL. To the best of our knowledge, \algname\ is the first deep offline RL algorithm specifically designed for MFGs, capable of handling arbitrarily sized datasets. This innovation opens the door for applying MFG theory in real-world systems where online data collection is prohibitive.
The primary challenge in adapting MFGs to the offline setting stems from the fact that these systems typically require estimating the distribution of agents (the mean-field), which complicates the direct adaptation of online algorithms. To address this, we repurpose ideas from off-policy policy evaluation (OPE) to approximate the mean-field distribution using offline data. Additionally, we apply a robust regularization mechanism to mitigate the effects of distributional shift, a well-known issue in offline RL where the learning policy encounters states or actions not sufficiently represented in the dataset. This regularization stabilizes the policy learning process, ensuring more reliable performance even in underrepresented areas of the state space.

We empirically validate \algname on a suite of benchmark tasks to demonstrate its efficacy. Specifically, we evaluate its performance on tasks involving both exploration and crowd navigation, two common challenges in the field of MFGs. Additionally, we explore the sensitivity of \algname to the quality of the datasets used for training, assessing how variations in state-action coverage and trajectory quality impact performance. We further investigate the importance of the proposed regularization term, designed to prevent the overestimation of $Q$-values --- a well-documented issue in offline RL algorithms.
In summary, our contributions are:
\begin{enumerate}
    \item \algnamelong, a novel deep RL algorithm for offline learning for MFGs.
    \item Extensive evaluation of the performance and ablation studies with respect to performance and dataset quality.
\end{enumerate}
\section{Background}\label{sec:background}
Sequential decision making problems commonly make use of finite horizon Markov decision process (MDP).
A finite horizon MDP is a tuple $\langle \mathcal{S}, \mathcal{A}, r, p, \gamma, H, \mu_0 \rangle$ consisting of a set of states $\mathcal{S}$, a set of actions $\mathcal{A}$, a reward function $r: \mathcal{S} \times \mathcal{A} \mapsto \mathbb{R}$, stochastic dynamics $p: \mathcal{S} \times \mathcal{A} \mapsto \Delta_\mathcal{S}$, discount factor $\gamma \in (0,1)$, horizon $H \in \mathbb{N}$ and initial state distribution $\mu_0 \in \Delta_\mathcal{S}$.
Problems involving a large number of interacting agents become intractable as the size of the state and action space grows exponentially with the number of agents. In many-agent games, where agents are anonymous and identical, \emph{Mean-Field Games} (MFGs) offer an effective framework to approximate Nash equilibria by simplifying the interactions between agents. 
First introduced by \citet{lasry2007mfg} and \citet{hang2006large}, MFGs address this complexity by modeling interactions through the distribution of agent states, rather than tracking individual agents. This reduces the dimensionality of the problem, making it more tractable and allows to approximate finite, $N$-player games.
In MFGs, a representative agent interacts with the mean-field (i.e., the distribution of all agents), rather than directly interacting with each individual agent. Consequently, the problem becomes optimizing a single policy with respect to this population distribution, which leads to more computationally efficient algorithms.

In this work, we consider \emph{stochastic, finite-horizon} MFGs with a finite set of states $\mathcal{S}$ and actions $\mathcal{A}$. The \emph{mean-field}, which is the distribution of agents over states at time $t$, is denoted by $\mu_t \in \Delta_{\mathcal{S}}$.
In the most general form, MFGs allow for mean-field-dependent dynamics, rewards, and even policies.
However, in this work, we focus on the case where only the rewards depend on the mean-field:
\begin{equation}
    r_t: \mathcal{S} \times \mathcal{A} \times \Delta_{\mathcal{S}} \mapsto \mathbb{R}.
\end{equation}

Given a policy $\pi$, the mean-field flow $\mu^\pi$ is defined by the following recursive relation, denoted by the mean-field evaluation operator $\phi(\pi) = \mu_\pi$:
\begin{align}
    \mu^\pi_{t+1}(s') &= \sum_{s \in \mathcal{S}} \sum_{a \in \mathcal{A}} p(s'|s,a) \pi(a|s) \mu^\pi_t(s), \\
    \mu^\pi_0 &= m_0, \label{eq:mf-flow}
\end{align}
where $p(s'|s,a)$ represents the transition dynamics of the environment, and $m_0$ is the initial distribution over states.
The goal for an agent is to find a policy $\pi: \mathcal{S} \mapsto \Delta_\mathcal{A}$ which maximizes the expected sum of rewards with respect to a given mean-field flow $\mu$:
\begin{gather}
    \max_\pi \quad J(\pi, \mu) = \mathbb{E}_\pi \Big[ \sum^{T}_{t=0} \gamma^t \, r(s_t, a_t, \mu_t) \Big]\nonumber\\
    \begin{aligned}
        \text{subject to:} \quad & s_0 \sim \mu_0 \nonumber\\
                    &  a_t \sim \pi(\cdot | s_t) \\
                    & s_{t+1} \sim p(\cdot | s_{t}, a_{t}).
    \end{aligned}
\end{gather}
By making the reward depend only on other agents via the mean-field, instead of the individual states and actions of all other agents, we obtain a much smaller optimization problem to solve. 

\paragraph{\textbf{Learning in MFGs}}
In contrast to single-agent RL, where we optimize a stationary reward signal, algorithms for MFGs typically aim to find policies that are close to some equilibrium concept because their performance largely depends on other agents.
The concept of Nash-Equilibria (NE), a common solution concept in game theory, has been extended to MFGs~\cite{lasry2007mfg} and is the main optimization target for algorithms solving non-cooperative MFGs. 

\begin{definition}
The best response (BR) to a mean-field flow $\mu$ is the solution of the optimization problem
$$\pi^* = \underset{\pi}{\text{argmax}} \, J(\pi, \mu)= BR(\mu). $$
\end{definition}

\begin{definition}[$\epsilon$-MFNE]
    A $\epsilon$-Mean-Field Nash Equilibrium with $\epsilon \geq 0$ is defined as a tuple $(\pi, \mu^\pi)$ for which the following holds:
    $$
    \underset{\pi' \in \Pi}{\text{sup}} J(\pi', \mu^\pi) \leq J(\pi, \mu^\pi) + \epsilon.
    $$
\end{definition}
Early methods to solve MFGs primarily involved solving coupled partial differential equations, typically a forward-backward system of Hamilton-Jacobi-Bellman and Fokker-Planck-Kolmogorov equations, to compute the value function and distribution flow of agents \cite{achdou2010mfg_numerical,achdou2012mfg_numerical}.
However, these methods struggle with scalability in high-dimensional state-action spaces and complex environments.
Algorithms for solving MFGs using RL commonly rely on some form of fixed point iteration,
alternating between policy updates and the mean-field distribution computation. 
Thereby, they use a best response computation step before computing the mean-field. 
On convergence, the fixed point iteration 
$$\phi(\pi^*) = \phi(BR(\mu^*)) = \mu^*$$ 
yields a MFNE.
However, generally convergence is not guaranteed~\cite{cui2021approximately_solving} and methods from algorithmic game theory, such as Fictitious Play (FP)~\cite{brown51fictitious}, are used to stabilize training. 
In recent years, machine learning approaches, particularly RL, have been explored as a promising alternative for solving MFGs \cite{yang2018mf_marl,guo2019learning, carmona2021modelfreemeanfieldreinforcementlearning}.
One of the central challenges in using RL to solve MFGs lies in the non-stationarity introduced by multi-agent interactions, which complicates the learning process.
Deep Learning variants of FP~\cite{pmlr-v37-heinrich15} were adapted to the MFG settings to scale to larger state and action spaces~\cite{cardaliaguet2017fictitious,xie2021learning, perrin2020fictitious,perrin2021flock}. 

In this work, we focus on a class of algorithms that evaluate a policy instead of computing a BR at each iteration~\cite{cacace2021policy_iteration}. Specifically, we adapt the Online Mirror Descent algorithm (OMD)~\cite{perolat2022omd,lauriere2022scalable} to the offline setting.
OMD  alternates between policy evluation and mean-field updates, as outlined in \Cref{alg:mirror_descent}.
The main difference to BR algorithms is that OMD tracks the sum of previous $Q$-values instead of policies.
The policy update in OMD is a softmax over the sum of previous $Q$-values.
However, it is not straightforward to sum up $Q$ functions in the case of nonlinear function approximators, such as neural networks.
This can be avoided by applying the \emph{Munchausen Trick}, as it was proposed in ~\cite{lauriere2022scalable}:
\begin{align}
    \pi^i &= \text{softmax} \Big(\frac{1}{\tau} \sum^i_{j=0} Q^j \Big) \\
     &= \text{argmax}_{\pi \in \Delta_\mathcal{A}}  \Big( \langle \pi, Q^i \rangle  - \tau \, \text{KL} \big(\pi||\pi^{i-1} \big) \Big) \\
     &= \text{argmax}_{\pi \in \Delta_\mathcal{A}}  \Big( \langle \pi, \underbrace{Q^i + \tau \, \ln{\pi^{i-1}}}_{\tilde{Q}^i} \rangle  - \tau \, \underbrace{\langle \pi, \ln{\pi} \rangle}_{\mathcal{H}(\pi)} \Big) \label{eq:munchausen_trick} \\
     &= \softmax \Big( \frac{1}{\tau} \, \tilde{Q}^i \Big).
\end{align}
This insight allows us to apply policy evaluation directly on $\tilde{Q}$ to avoid the summation of $Q$-functions.
In \Cref{eq:munchausen_trick}, $\langle \pi, Q^i \rangle$ denotes the shorthand notation for $\sum_a \pi(a|s) \, Q^i(s,a)$.
The modified Bellman Operator is then defined as:
\begin{align}
    (\mathcal{B}_\mu^\pi \, \tilde{Q})(s,a) &= \tilde{r} + \gamma \, \mathbb{E}_{s',a'} [ \tilde{Q}(s',a') - \tau \, \ln{\pi^{i-1}(a'|s')} ] \label{eq:momd_operator}\\
    \tilde{r}(s,a,\mu) &= r(s,a,\mu) + \tau \, \alpha \ln{\pi^{i-1}(a|s)},
\end{align}
with modified reward $\tilde{r}$, which penalizes deviations from the previous policy $\pi^{i-1}$.
The hyperparameter $\tau$ acts as a \emph{temperature} and scales the sum of $Q$-values to avoid premature convergence whereas $\alpha$ is a regularization parameter to control how far a new policy can be from the previous policy.
For a more detailed derivation, we refer to~\cite{lauriere2022scalable}.

\begin{algorithm}
\caption{Munchausen Online Mirror Descent for MFGs~\cite{lauriere2022scalable}}
\begin{algorithmic}[1]
     \For{$i$ = 1 ... $L$}
        \State \textbf{Mean-Field Update:} $\mu^{i} \leftarrow \phi(\pi^{i}) $
        \State \textbf{Regularized Policy Evaluation:} $\tilde{Q}^{i+1} \leftarrow \mathcal{B}^{\pi^{i}}_{\mu^{i}} \tilde{Q}^{i}$
        \State \textbf{Policy Update:} $\pi^{i+1}(\cdot|s) \leftarrow \text{softmax}(\frac{1}{\tau}\tilde{Q}^{i+1}(s,\cdot))$     
    \EndFor
\end{algorithmic}\label{alg:mirror_descent}
\end{algorithm}

\section{Related Work}

Recent years brought up many works that address various limitations of MFGs. 
\citet{lauriere2022scalable} introduced the basis of our work, Deep Munchausen Mirror Descent, a deep neural network based variant of OMD~\cite{perolat2022omd} with strong empirical performance.
Also \citet{cui2021approximately_solving} and \citet{perrin2022scaling} introduce deep RL based approaches on MFGs.
Other works address the assumption of identical agents and extend MFGs to multi-population games~\cite{carmona2019stochasticgraphongamesi,fabian2023graphon}. 
\citet{subramanian2022decentralized} introduce decentralized MFGs, allowing to lift the assumption of indistinguishable agents.
Inverse-RL (IRL) methods have been applied to the MFG setting to infer unknown reward signals~\cite{chen2022mfg-irl,chen2023mfg-irl}.
\citet{yang2018irl} introduce an IRL approach that learns the dynamics and the reward model from data. However, their work focuses on behavior prediction rather than finding MFNE.
Recent work on model-based algorithms in the mean-field control (MFC) setting, a subclass of MFGs in which agents fully cooperate, can learn a model of the environment and use that to optimize a MFC policy with better sample efficiency~\cite{pasztor2023efficient,huang2024statistical}. \citet{matej2024safe} extend this to safety-constrained problems.
However, although those approaches learn a model of the environment, they still assume access to the environment to generate new data for exploration.

Despite many advances in the field of MFGs, the direction of purely offline learning remains an underexplored topic.
SAFARI~\cite{chen2021pessimism} is, to our knowledge, the only approach specifically designed for offline mean-field RL, which is not directly comparable to our approach as it does not approximate MFNE.
Its main innovation lies in using RKHS (Reproducing Kernel Hilbert Space) embeddings to model the mean-field distribution, combined with an uncertainty-regularized value iteration based on fixed trajectories. While theoretically robust, with dataset-dependent performance bounds, SAFARI faces significant scalability issues. The RKHS embeddings require inverting a Gram matrix that grows quadratically with the dataset size, leading to substantial memory demands and computational bottlenecks.
\section{Method}
In this section we discuss the foundational components of \algname.
In particular, we elaborate how we can leverage methods from offline policy evaluation to estimate the mean-field $\mu$ from static datasets and provide a modified version of the D-MOMD algorithm to adapt it to the offline learning regime.

\subsection{Offline Mean-Field Estimation}\label{sec:mean_field_estimation}
Fixed point algorithms for solving mean-field games, as discussed in \Cref{sec:background}, iterate between (1) evaluating the mean-field distribution of agents and (2) best-response computation or policy evaluation. 
The first step, mean-field estimation, can be done via direct computation if one has access to the transition model or via Monte-Carlo samples if a simulator is provided. 
In scenarios where only a static dataset of previously collected environment interactions are available, online algorithms like D-MOMD~\cite{lauriere2022scalable} are not applicable.

Therefore, we seek to approximate the mean-field flow
\begin{align}
    \mu^\pi_{t+1}(s') &= \sum_{s \in S} \sum_{a \in A} \, p(s'|s,a) \, \pi(a|s) \, \mu^\pi_t(s)
\end{align}
without having access to the transition model $p(s'|s,a)$.
One approach to mitigate this is to leverage the data to learn a model of the dynamics.
However, the models may be inaccurate in large action
spaces, where not all actions are frequently visited. Moreover, approximating the environment dynamics with neural networks might cause additional biases from covariate shifts due to the change of policies~\cite{xie2019mis}.
In this work, we leverage the fact that this problem can be equivalently treated as an off-policy state density estimation problem.
OPE methods are designed to estimate quantities such as rewards or value functions from off-policy samples. 
We repurpose this idea to estimate the state distribution under a new policy.
In particular, we are interested to estimate $\mu_\pi(s)$ from samples that are not collected from $\pi$ but some other, possibly unknown, behavior policy $\pi_\beta$ and without having access to environment dynamics $p(s'|s,a)$.

Let $d_\mu^\pi(s,a) = \pi(a|s)\,\mu_\pi(s)$ be the joint state-action distribution given a mean-field $\mu_\pi$ and policy $\pi$.
We restate the mean-field flow as an expectation over $d_\mu^\pi$:
\begin{align}
    \mu_\pi^{t+1}(s') &= \mathbb{E}_{(s,a) \sim d_{\mu^t}^\pi} \Big[ p(s'|s,a) \Big].
    \label{eq:exp_mf}
\end{align}
In general, we could apply any OPE method capable of estimating density (ratios) such as model-based estimators~\cite{kidambi2020morel,yu2020mopo} or DICE-style approaches for estimating stationary distributions~\cite{wen2020batch_stationary_distribution,lee2021optidice}, as \algname is agnostic to the estimation method.
In this work we decide for using importance sampling to estimate $\mu_\pi^{t+1}$. In particular, we make use of the marginalized importance sampling (MIS) estimator of \citet{xie2019mis} because of its theoretical properties and its simplicity compared to other approachs, which typically require additional steps, such as solving an inner optimization problem~\cite{nachum2020dualdice,zhang2020GenDICE} or fitting another model~\cite{yu2020mopo,kidambi2020morel}.

Let $d(s,a)=\pi_\beta(a|s) \, d(s)$ be the joint state-action distribution of the dataset collected under behavior policy $\pi_\beta$.
In practice, $\pi_\beta$ is often unknown and can be approximated using the state-conditional empirical distribution over actions in $\mathcal{D}$~\cite{kumar2020cql}.
%$$\hat{\pi}_\beta(s,a) = \frac{\sum_{s',a' \in \mathcal{D}} %\mathbbm{1}[s=s', a=a']}{\sum_{s' \in \mathcal{D}} %\mathbbm{1}[s=s']}.
%$$
We can apply importance sampling to reformulate \Cref{eq:exp_mf} as
\begin{align}
    \mu_\pi^{t+1}(s') &= \mathbb{E}_{(s,a) \sim d} \Bigg[  \frac{d_{\mu^t}^\pi(s,a)}{d(s,a)} \, p(s'|s,a) \Bigg] \\
    &= \mathbb{E}_{(s,a) \sim d} \Big[  \frac{\pi(a|s)\,\mu^t_\pi(s)}{\pi_\beta(a|s)\,d(s)} \, p(s'|s,a) \Big].\label{eq:mf_importance_sampling}
\end{align}
This factorization of the state-action distribution allows us to apply MIS to approximate \Cref{eq:mf_importance_sampling} using samples $(s^i_t,a^i_t,s^i_{t+1})$ from finite dataset $\mathcal{D}$.
Let $\hat{d}(s_t) = \frac{1}{|\mathcal{D}|} \sum_i^{|\mathcal{D}|} \mathbbm{1}[s_t^{(i)}=s_t]$ denote the empirical state distribution at time $t$, then the marginalized state distribution can be estimated recursively by
\begin{equation}
\begin{split}    
    \mu_\pi^{t+1}(s) &\approx \frac{1}{|\mathcal{D}|} \sum^{|\mathcal{D}|}_{i=0} \frac{\mu_\pi^t(s_t^{(i)})}{\hat{d}(s_t^{(i)})} \, \frac{\pi(a_t^{(i)}|s_t^{(i)})}{\pi^\beta(a_t^{(i)}|s_t^{(i)})} \, \mathbbm{1}[ \ s^{(i)}_{t+1} = s ] \\
    \mu_\pi^0(s) &= \hat{d}(s_0). \label{eq:mis}
    \end{split}
\end{equation}
This yields an unbiased estimator of $\mu_\pi$ with polynomial error bound with respect to time horizon $H$, which reduces to $\mathcal{O}(H)$ in some cases, such as bounded maximum expected returns~\cite{xie2019mis}.
Note that, unlike in typical single-agent offline RL scenarios, we can not directly estimate reward $r_t$, as it depends nonlinearly on $\mu_t$ in general. 
Thus, for the general case, we require access to the reward function. 
For special cases, such as reward functions monotonic in $\mu$, we could in principle approximate $r_t$ directly.

\subsection{Offline Munchausen Mirror Descent}

In \Cref{sec:mean_field_estimation}, we introduced an offline method for estimating the mean-field distribution of a policy. This method can, in principle, be directly applied to D-MOMD to adapt it to the offline learning setting. 
The update rule in our algorithm follows the classic TD-error minimization approach, as used in DQN~\cite{mnih2015dqn}, where the target $Q$-values are parameterized by $\bar{\theta}$. Specifically, the objective is to minimize the temporal-difference error where the policy evaluation operator is defined as in \Cref{eq:momd_operator}:
\begin{equation}
    \min_\theta \mathbb{E}_{(s,a,s') \sim \mathcal{D}} \Big[ \big( Q_\theta(s,a) - (\mathcal{B}_\mu^\pi Q_{\bar{\theta}})(s, a) \big)^2 \Big].
\end{equation}

However, naively applying off-policy algorithms to offline RL tasks typically leads to the overestimation of $Q$-values for actions not well represented in the dataset. To address this issue, we incorporate a regularization term following the idea of \emph{Conservative Q-Learning} (CQL)~\cite{kumar2020cql}, which is designed to learn a conservative lower bound of the true $Q$-function.
CQL introduces a regularized version of the Bellman equation, where the objective balances the maximization of the $Q$-values over the dataset and the minimization of the temporal-difference error. The general CQL optimization problem is given as:
\begin{equation}
\begin{split}
    \min_Q \max_{\tilde{\pi}} \, \mathbb{E}_{(s,a) \sim \mathcal{D}, \tilde{\pi}} \big[ Q(s,a) \big] &- \mathbb{E}_{(s,a) \sim \mathcal{D}, \pi_\beta} \big[ Q(s,a) \big] \\
    &+ \big| Q - \mathcal{B}^* Q \big|^2 + \mathcal{R}(\tilde{\pi}), \label{eq:cql}
\end{split}
\end{equation}
where $\tilde{\pi}$ represents a policy used to define the joint-action distribution over which we minimize the state-action values. The second term encourages tighter bounds by maximizing $Q$-values under the dataset distribution. The last term is the classic Bellman equation minimizing the TD error with a regularization term $\mathcal{R}$ applied to $\tilde{\pi}$.
For specific choices of $\mathcal{R}$, the inner maximization problem can be solved in closed form. A common choice for $\mathcal{R}$ is to use the KL divergence to some prior action distribution $\rho$. If we chose $\rho$ to be the uniform distribution over actions, we obtain an entropy regularized, closed-form loss function for \algname:
\begin{equation}
\begin{split}
    \mathcal{L}(\theta) = \, & \eta \, \mathbb{E}_{(s,a) \sim \mathcal{D}} \Big[\log{\sum_{a'} \exp{Q_\theta(s,a')}} - Q_\theta(s,a) \Big] \\
    + & \,  \mathbb{E}_{(s,a,s') \sim \mathcal{D}} \Big[ \big( Q_\theta(s,a) - (\mathcal{B}_\mu^\pi Q_{\bar{\theta}})(s,a) \big)^2 \Big], \label{eq:loss}
\end{split}
\end{equation}
where $\eta$ is a hyperparameter to control the importance of the regularization.

The pseudo-code for \algname is shown in \Cref{alg:offmd}. We use \Cref{eq:mis} to compute the offline mean-field and \Cref{eq:loss} to compute the loss function and update the parameters $\theta$ via stochastic gradient descent.
\algname thus follows the same iterative schema as \Cref{alg:mirror_descent}.

\begin{algorithm}
\caption{Offline Munchausen Mirror Descent (\algname)}
\begin{algorithmic}[1]
    \State \textbf{Input:} Dataset $\mathcal{D}$, initial parameters $\theta^1$
    \For{$i$ = 1 ... $L$}
     \State Estimate mean-field $\mu^{i}$ using \cref{eq:mis}
        \For{$j$ = 1 ... $B$}
            \State Sample batch $\mathcal{B}$: $\{(s^k_t,a^k_t,s^k_{t+1})\}_{k=1}^N \sim \mathcal{D}$
            \State Relabel reward using $\mu^i_t$: $r^k_t = r(s^k_t,a^k_t, \mu_t^{i}$)
            \State Update: $\theta_i \leftarrow \theta_i - \nabla_\theta \mathcal{L}(\theta_i)$  using \cref{eq:loss}
    \EndFor
    \State $\theta^{i+1} \leftarrow \theta^i$
    \State Update policy: $\pi(a|s) = \text{softmax}\big({\frac{1}{\tau} \bar{Q}_{\theta_{i+1}}(s,a)}\big)$
\EndFor
\end{algorithmic}\label{alg:offmd}
\end{algorithm}

\section{Experimental Evaluation}

We evaluate the Off-MOMD algorithm on two grid-world problems introduced by \citet{lauriere2022scalable} and compare its performance against the online variant.
We also conduct experiments to investigate the sensitivity to the quality of the dataset and investigate the importance of the conservative constraint on the loss function.
The algorithms and the environments are implemented in JAX~\cite{jax2018github} and build on code by \citet{jaxrl} and \citet{lanctot2019openspiel}.
The code for the experiments is available on GitHub.\footnote{\url{https://github.com/axelbr/offline-mmd}}

\subsection{Experiment Setup}

To make runs comparable with each other, we employ \emph{Exploitability} as an evaluation criteria (also often referred to as \emph{Regret}):
$$
\mathcal{E}(\pi, \mu) = \max_{\pi'} J(\pi', \mu) - J(\pi, \mu). 
$$
It directly measures how far a learned policy is from a MFNE by quantifying the potential utility an agent can gain by deviating from its policy, with lower exploitability indicating better equilibrium approximation.
We use the same evaluation protocol as in~\cite{lauriere2022scalable} and compute the ground truth mean-field and exploitability.

Both, \algname and D-MOMD, optimize a $Q$ function represented as a neural network with 3 layers of 128 parameters and ReLU activations.
The hyperparamer settings are the same for all instances of \algname over all tasks, except the ablation studies.
For more details, we refer to \Cref{sec:appendix}.

\subsection{Performance Evaluation}

We evaluate the performance of our algorithm on two distinct tasks within a gridworld environment consisting of four separated rooms connected by narrow corridors, as described in \cite{lauriere2022scalable,lauriere2024learningmeanfieldgames}. Agents can choose from five actions: move up, down, left, right, or stay in place. If an action results in a collision with a wall, the agent remains in its current position. The time horizon for each episode is set to 40 timesteps. The two tasks we evaluate are exploration and crowd navigation.
For each task, we train Off-MMD on three datasets of varying quality and compare its exploitability against the baseline. The following variants of Off-MMD are included in the evaluation:

\begin{itemize}
    \item \textbf{D-MOMD}: The online baseline algorithm.
    \item \textbf{\algname (Exp)}: Trained on data collected by a fully trained D-MOMD policy.
    \item \textbf{\algname (Int)}: Trained on data collected from an intermediate checkpoint.
    \item \textbf{\algname (Rand)}: Trained on data collected from a uniform random policy.
\end{itemize}
All datasets contain 100K episodes with 40 timesteps each.
The subsequent sections present the evaluation results for the exploration and crowd navigation tasks.

\paragraph{\textbf{Exploration Task}}
In this task, agents start in the left upper corner and must spread evenly across all four rooms. The reward function is defined as 
$$r(s_t, a_t, \mu_t) = -\log{\mu_t(s_t)},$$
which incentivizes agents to occupy less crowded states, leading to higher rewards when low-density states are reached. 
The optimal policy for this task spreads evenly across the whole state-space.

\begin{figure}
     \centering
     \begin{subfigure}{\columnwidth}
         \centering
         \includegraphics[width=\columnwidth]{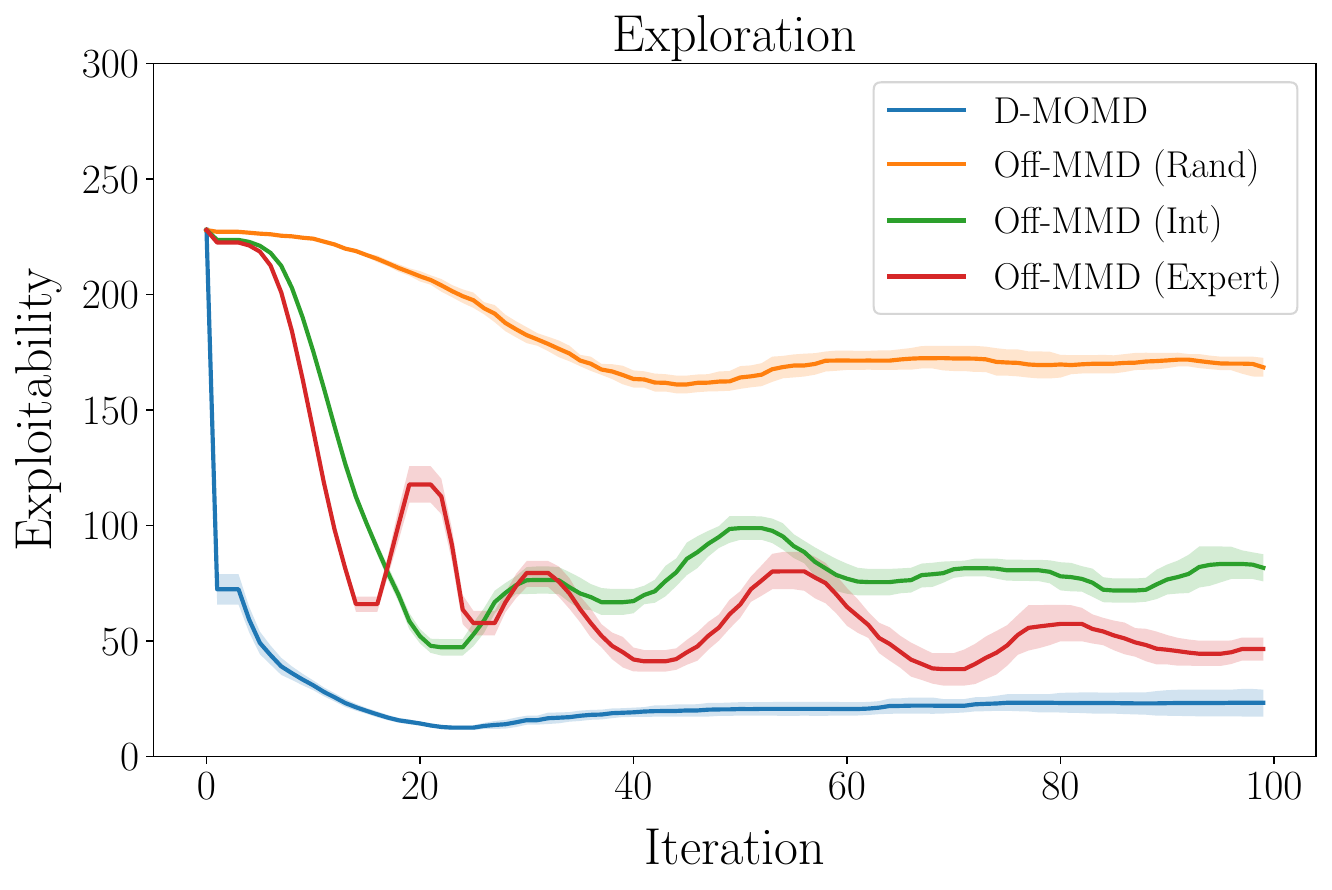}
         \caption{Exploitability}
         \label{fig:exploration_exploitablity}
     \end{subfigure}
     \hfill
     \begin{subfigure}[b]{\columnwidth}
         \centering
         \includegraphics[width=\columnwidth]{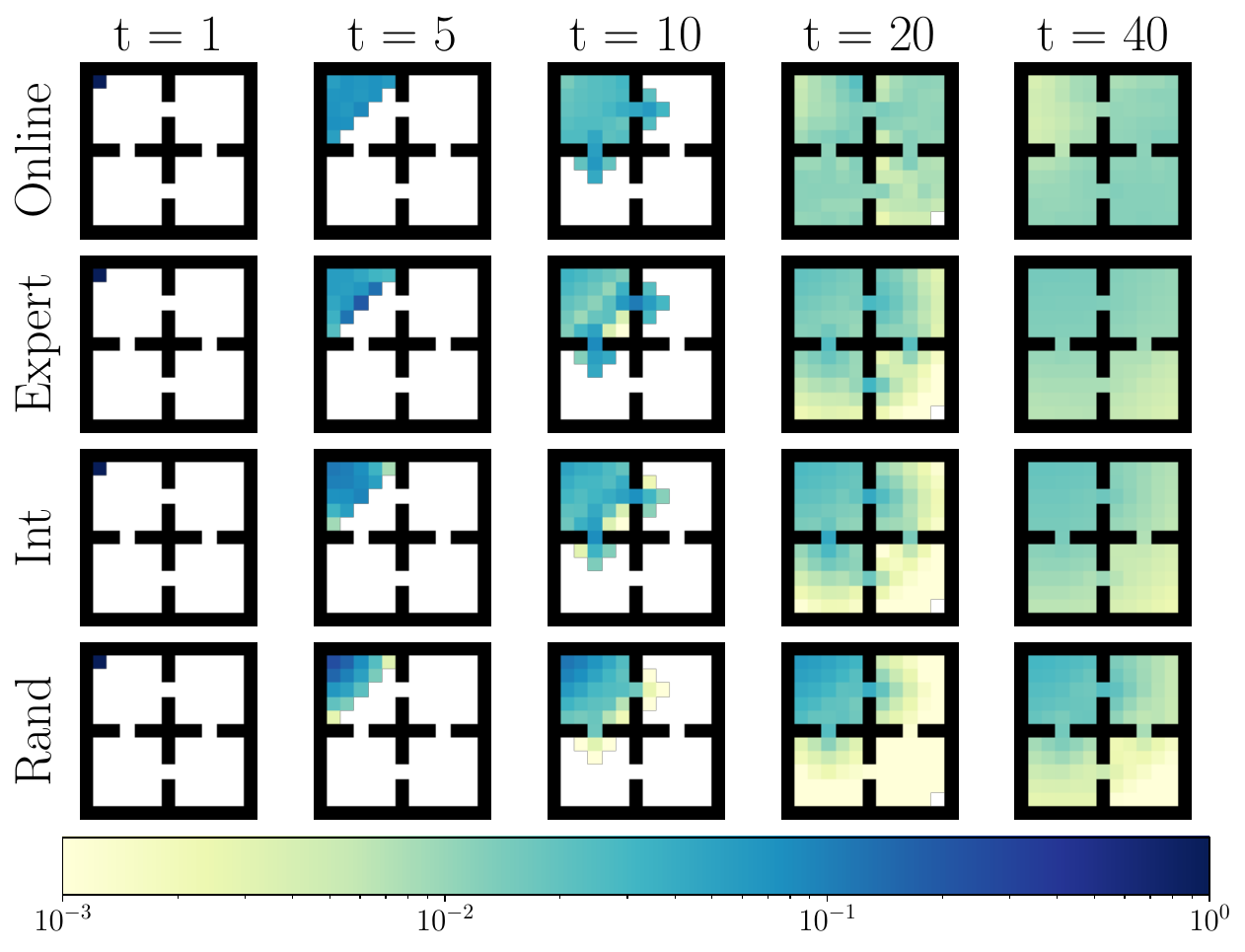}
         \caption{Mean-Field Evolution}
         \label{fig:exploration_evolution}
     \end{subfigure}
    \caption{(a) \algname can approximate the performance of D-MOMD on the Exploration task when being trained on reasonably good datasets. Training runs were conducted over 10 seeds for 100 iterations of \algname and D-MOMD. We report the mean exploitability and the 95\% confidence interval. (b) Evolution of the mean-field over timesteps $t$. Darker areas indicate higher state density.}
    \label{fig:exploration}
\end{figure}

In \Cref{fig:exploration_exploitablity}, we present the exploitability over 100 iterations of our algorithm. The online variant rapidly converges to the expected outcome. We evaluate the performance of three instances of \algname, each trained on a dataset of different quality. When trained on sufficiently high-quality datasets, \algname consistently learns policies that perform well. \Cref{fig:exploration_evolution} illustrates the evolution of the mean-field over time, supporting this observation. 
As expected, the policy trained on data generated by a uniform random policy fails to spread evenly across all rooms, particularly in the lower-right room, due to insufficient coverage of this region in the dataset.

\paragraph{\textbf{Crowd Modelling with Congestion}}

\begin{figure}
     \centering
     \begin{subfigure}[b]{\columnwidth}
         \centering
         \includegraphics[width=\textwidth]{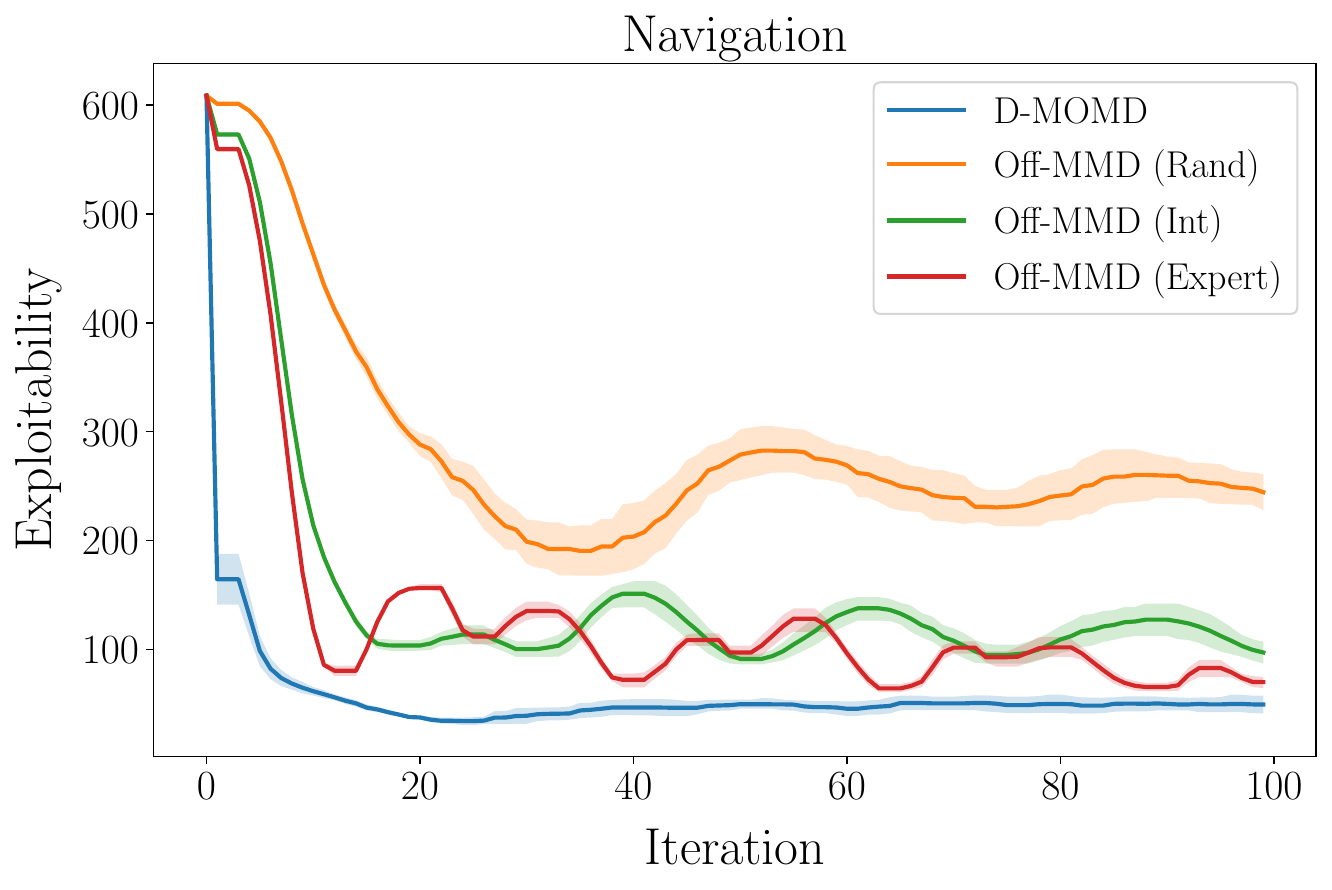}
         \caption{Exploitability}
         \label{fig:navigation_exploitability}
    
     \end{subfigure}
     \hfill
     \begin{subfigure}[b]{\columnwidth}
         \centering
         \includegraphics[width=\linewidth]{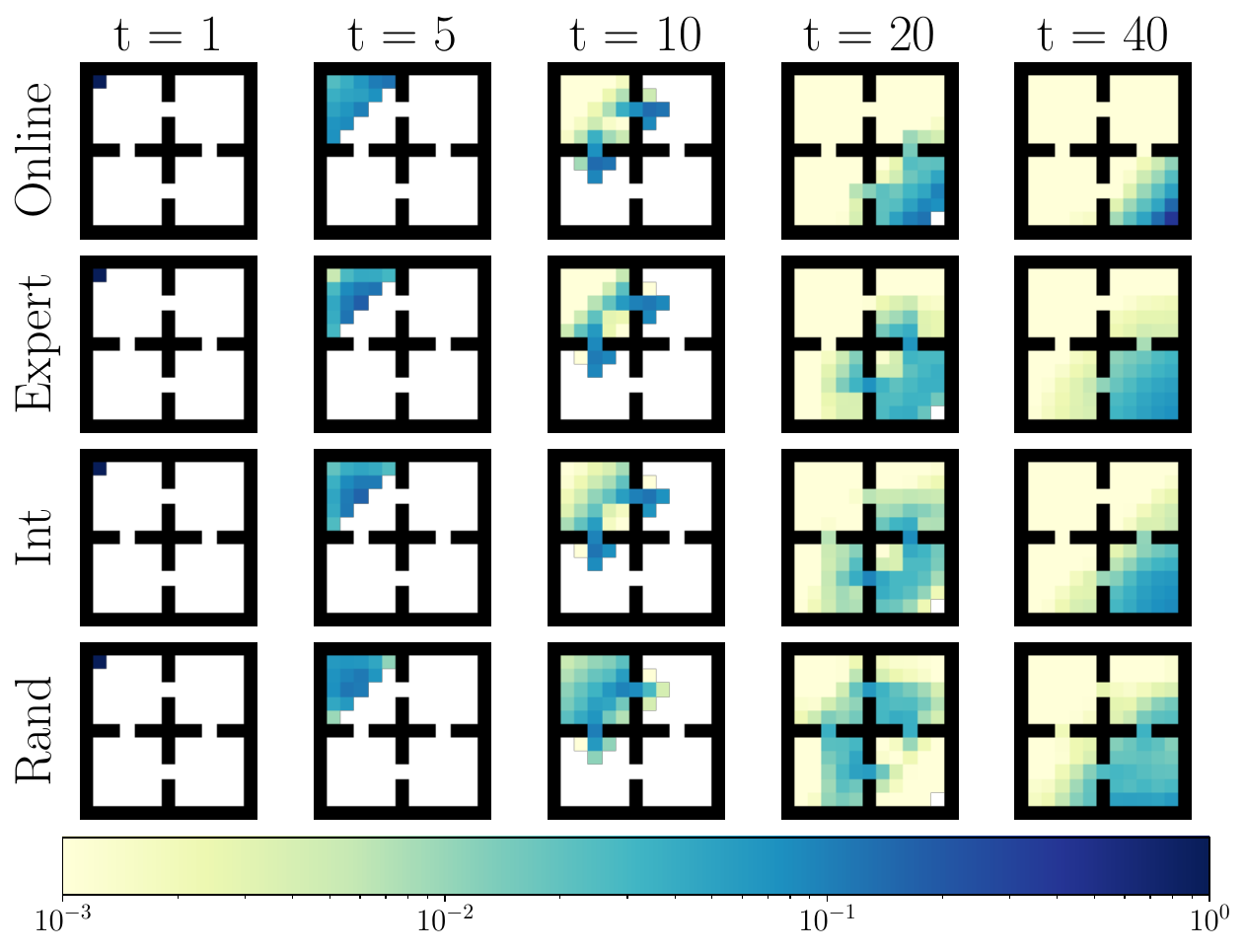}
         \caption{Mean-Field Evolution}
         \label{fig:navigation_evolution}
     \end{subfigure}
    \caption{\algname performs best with intermediate and expert quality datasets. Compared to the exploration task, the policy trained on the random behavior dataset performs better. Experiment settings are the same as in \Cref{fig:exploration}.}
    \label{fig:navigation}
\end{figure}

In this task, agents also start in the upper left corner. 
Differently to the exploration task, agents must navigate to the target position in the lower right corner while avoiding high-density areas. 
Furthermore, we simulate congestion effects by penalizing movements when agents are in crowded areas.
The reward function is defined as
$$
r(s_t, a_t, \mu_t) = -||s_t - s_{\text{target}}|| - \mu_t(s_t) \, ||a|| - \log{\mu_t(s_t)}.
$$
This is a more complex reward function, as it poses a trade-off of conflicting goals for the agents.
The results, shown in \Cref{fig:navigation_exploitability}, demonstrate convergence of \algname (Exp) and \algname (Int) towards the baseline. Notably, the policy trained on the random dataset also performs reasonably well. We hypothesize that the distance penalty provides effective guidance, enabling the policy to solve the task even with limited state-action coverage in certain parts of the state space.

\subsection{Impact of Dataset Quality}
In offline RL, we are interested in the robustness of policy performances to dataset quality.
In this experiment, we aim to investigate the sensitivity of \algname\ to changes in the quality of the trajectories in the dataset and the coverage of the state space.
Building on the methodology of \citet{schweighofer2022dataser}, we categorize datasets based on two criteria: \textit{state-action coverage} and \textit{trajectory quality}.
\begin{definition}[State-Action Coverage]
\label{def:cov}
    Let $u_{s,a}(\mathcal{D})$ denote the number of unique state-action pairs in a dataset $\mathcal{D}$, then the state-action coverage of this dataset is defined as
    \begin{equation}
    \text{Coverage}(\mathcal{D}) = \frac{u_{s,a}(\mathcal{D})}{|S||A|}.
    \end{equation}
\end{definition}
\begin{definition}[Trajectory Quality]\label{def:tq}
Let $g(\mathcal{D})$ denote the average episode return of a dataset. Furthermore, let $\mathcal{D}_{\text{min}}$ and $\mathcal{D}_{\text{expert}}$ be reference datasets, collected by a suboptimal and an expert policy, respectively. The trajectory quality of another dataset  $\mathcal{D}$ is defined as
    \begin{equation}
     \text{Quality}(\mathcal{D}) = \frac{g(\mathcal{D})-g(\mathcal{D}_{\text{min}})}{g(\mathcal{D}_{\text{expert}})-g(\mathcal{D}_{\text{min}})}.\label{eq:quality}
    \end{equation}
\end{definition}
In our experiments, we use datasets collected by the expert policy and the random policy for computing the normalization bounds in \Cref{eq:quality}.
Using the three datasets introduced previously for the navigation task, we generate 100 synthetic datasets by randomly subsampling episodes. The dataset sizes range from 1,000 to 100,000 episodes. We then train policies using \algname\ on these datasets to evaluate the effect of dataset quality on performance.

\begin{figure}
    \centering
    \includegraphics[width=\columnwidth]{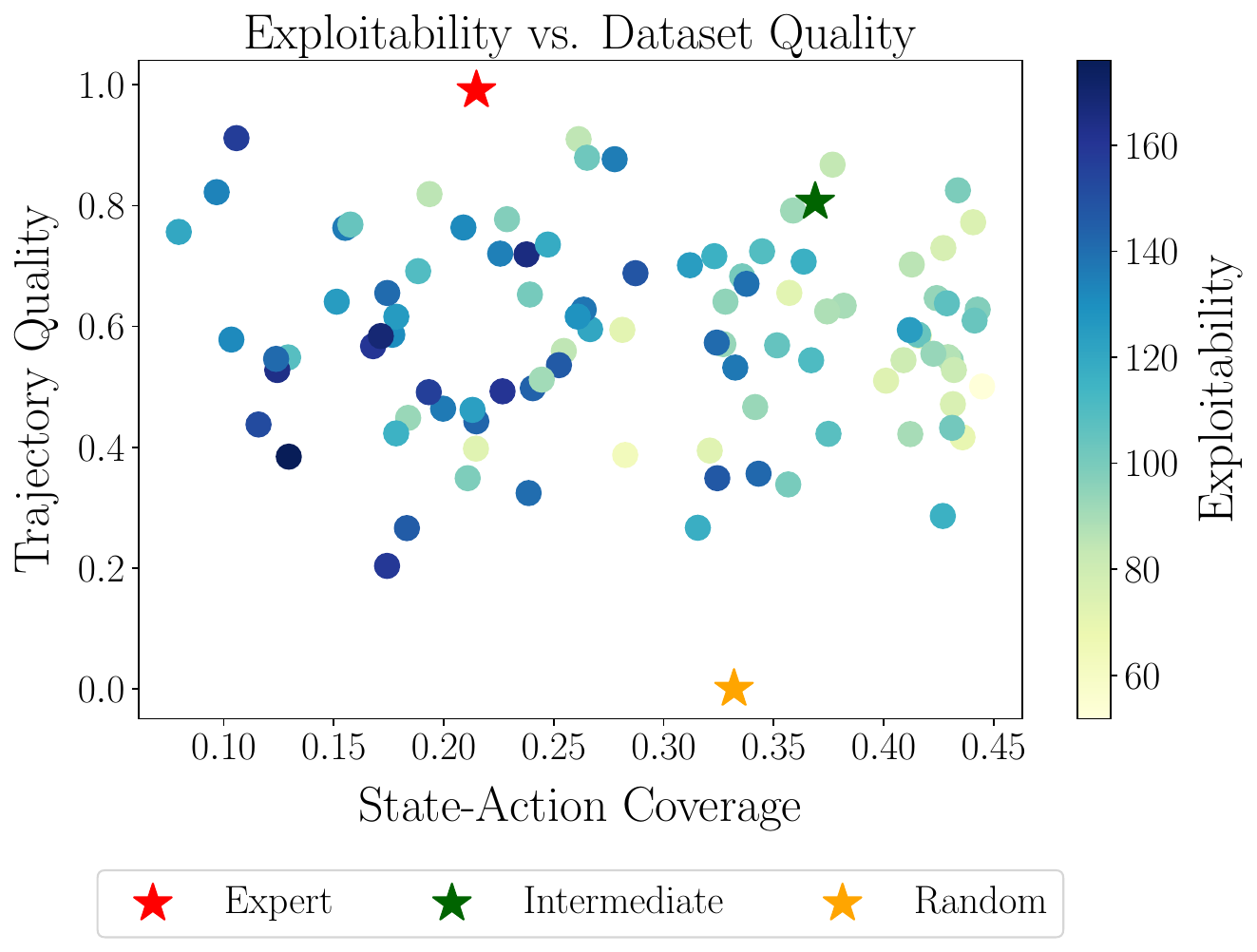}
    \caption{Exploitability vs. Data-Quality: Each point represents a training run of \algname on a dataset with specific state-action coverage and trajectory quality. The color indicates the exploitability of the policy after 100 iterations with darker colors indicating higher exploitability. For reference, we also mark the datasets used in previous experiments.}
    \label{fig:dataset_quality}

\end{figure}

\Cref{fig:dataset_quality} presents the exploitability of \algname\ when trained on datasets with varying state-action coverage and trajectory quality. The results indicate a strong correlation between state-action coverage and performance, whereas trajectory quality appears to be a weaker predictor, except in extreme cases such as expert demonstrations or fully random datasets (highlighted in \Cref{fig:dataset_quality}).
This behavior can be explained by the challenges inherent in multi-agent systems: the policy return is strongly influenced by the behavior of other agents. Therefore, performance achieved under one mean-field setting may not be comparable to another.

\begin{figure}
    \centering
    \includegraphics[width=\columnwidth]{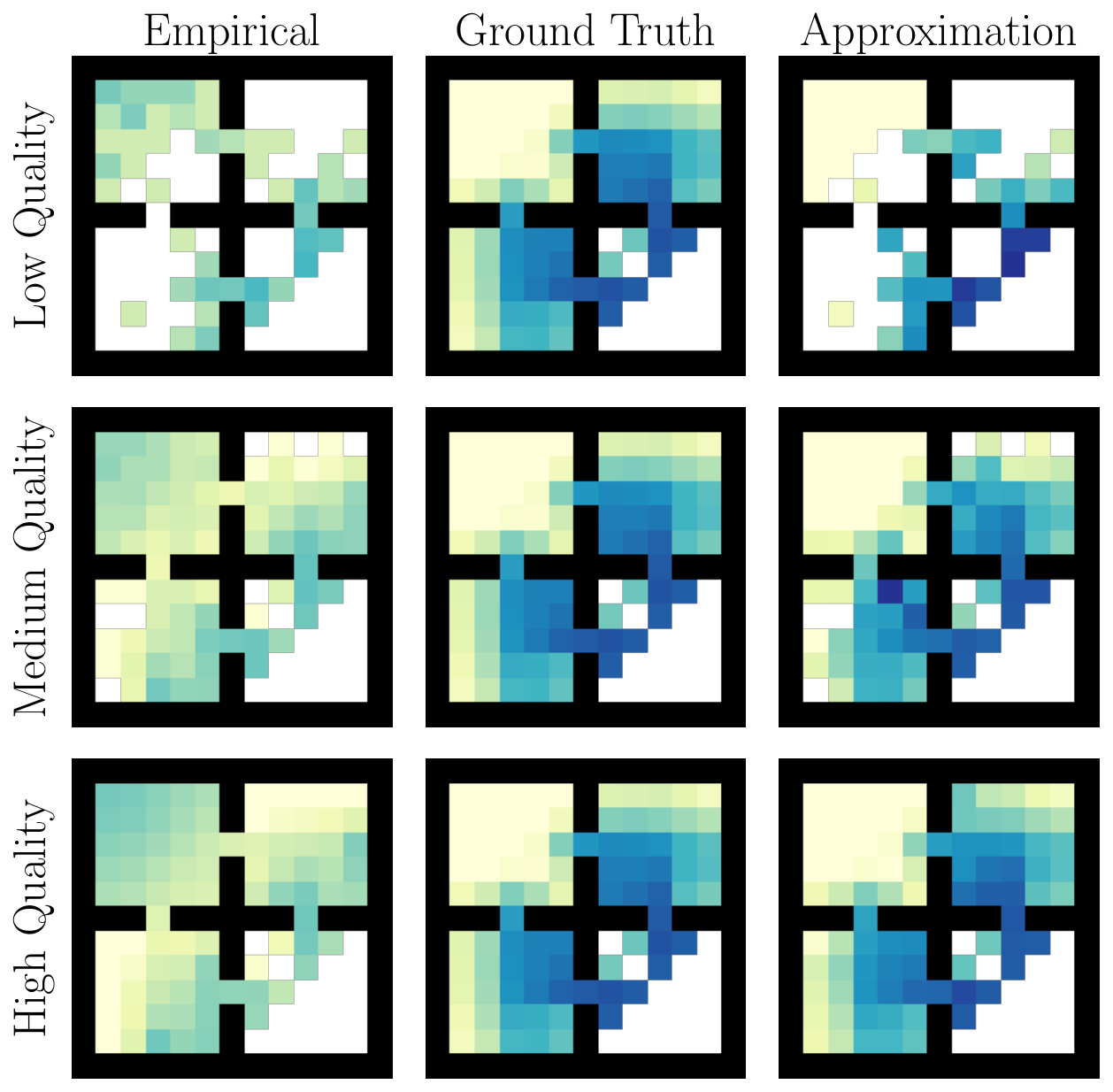}
    \caption{The left column shows the empirical state distribution of datasets collected by behavior policy $\pi^\beta$ (from the navigation task) with different state coverages. The center column shows the ground-truth mean-field that is generated by the new policy to evaluate and the right column shows the approximated mean-field of that policy using just the dataset. The mean-fields are picked at $t=15$. White spots indicate no state-action coverage in this area.}
    \label{fig:mf_approximation}

\end{figure}

In \Cref{fig:mf_approximation}, we visualize the approximation of the mean-field under datasets of varying quality for a fixed policy.
The leftmost column shows the empirical state distribution of the datasets whereas the center and right column shows the mean-field under some fixed policy. 
The ground truth mean-field serves as a reference for the approximation in the right-most column and is therefore independent of the dataset quality. 
\Cref{fig:mf_approximation} shows how the approximation of the mean-field changes with the state coverage in the dataset. 
This is particularly prevalent in the first row, where we chose the dataset with the lowest state coverage from the set of synthetic datasets. The approximation can not provide estimates for unvisited parts of the state space.
However, for the states that are in the dataset, it produces correct estimations of the mean-field distribution.
In scenarios with higher quality datasets, we observe accurate approximations of the ground-truth distribution.

\subsection{Effect of Regularization}
In the following, we investigate the importance of the CQL regularization term with respect to the robustness to varying levels of state-coverage in the dataset and the training stability of \algname.

\begin{figure}
    \centering
    \includegraphics[width=\columnwidth]{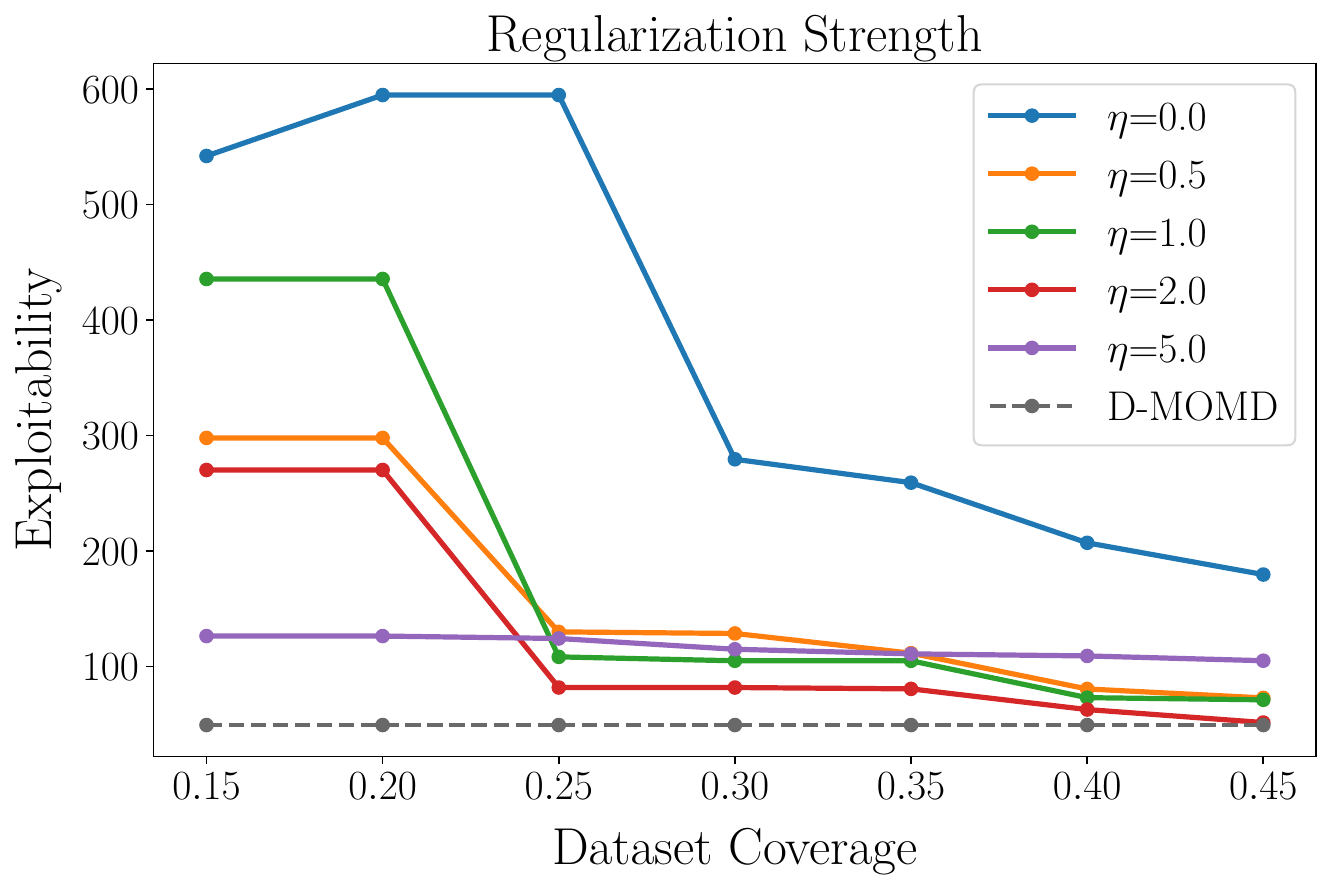}
    \caption{We report the exploitability of policies with different values of $\eta$ in \cref{eq:loss}. We train each configuration on 5 datasets that have state-action coverages close to a specific value, ranging from 0.15 to 0.45. We report the mean exploitability of the policies after 100 iterations. For reference, we also plot the performance of the online baseline after 100 iterations.}
    \label{fig:cql_ablation}
\end{figure}

The first experiment, shown in \Cref{fig:cql_ablation}, aims to examine the effect of the regularization hyperparameter $\eta$ on the quality of the policy.
We conduct training runs with varying regularization strengths, ranging from 0 (e.g. no regularization) to 5 across datasets with different levels of state-action coverage.
Specifically, we select five datasets closest to each state-action coverage bin, with coverage values ranging from 0.15 to 0.45. 
\algname\ is trained for 100 iterations on each of these datasets, allowing us to analyze the influence of regularization under diverse coverage conditions.
\Cref{fig:cql_ablation} shows that moderate regularization 
allows to reach comparable exploitability as D-MOMD on higher state-action coverages while being significantly more robust than \algname without regularization.
The best final performance is achieved with $\eta=2.0$, which coincides with the recommended setting for CQL~\cite{kumar2020cql}.
Furthermore, we can see that larger values for the regularization hyperparameter $\eta$ dampen the difference in performance over different coverage levels, but hurt the maximum achievable performance.

\begin{figure}
    \centering
    \includegraphics[width=\columnwidth]{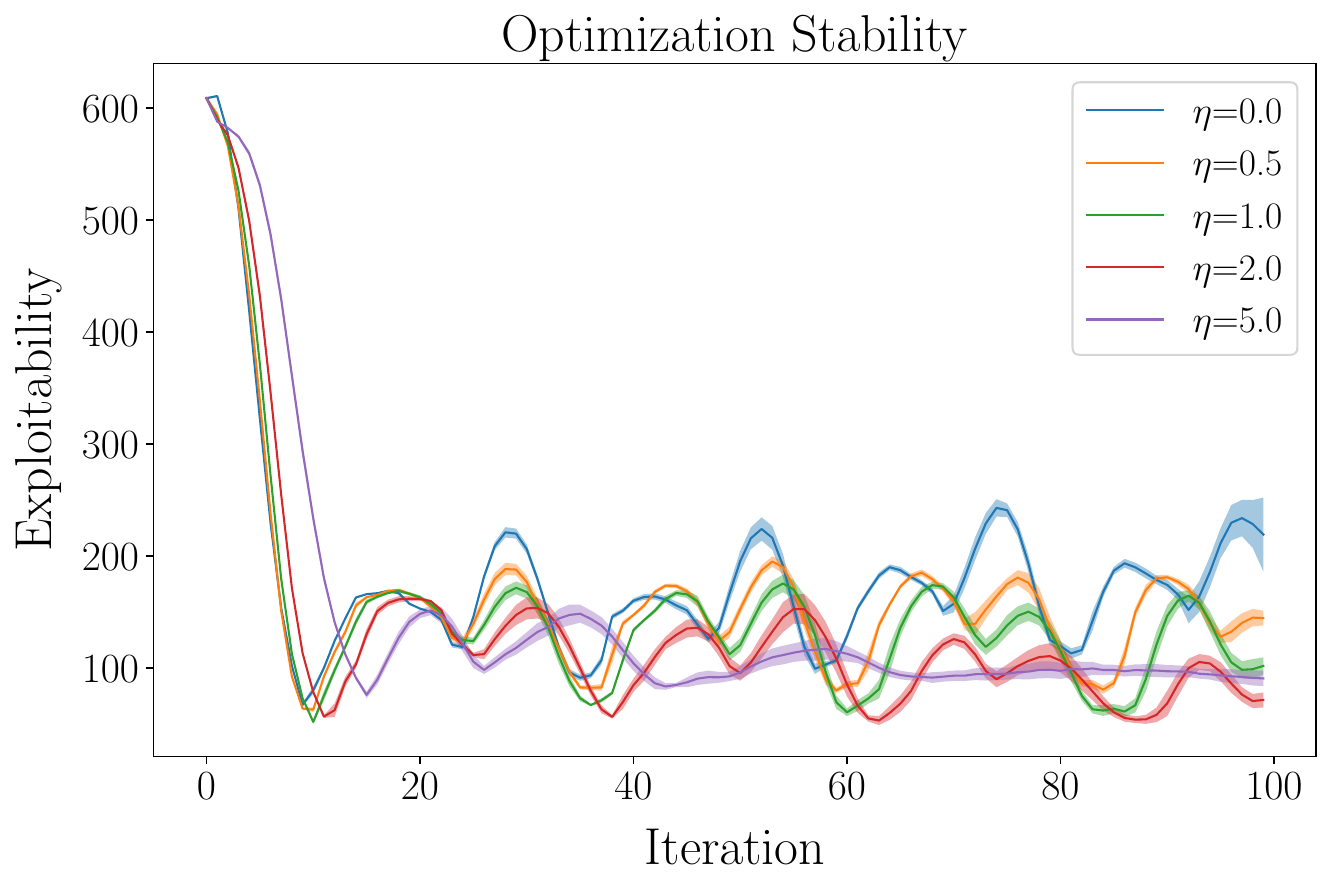}
    \caption{We optimize policies with different values of $\eta$ on the same expert dataset of the navigation tasks and plot the exploitability over training iterations. We show the mean and 95\% confidence interval over 10 seeds.}
    \label{fig:training_stability}
\end{figure}

In another ablation experiment, we investigate the training stability of \algname, specifically focusing on monotonic improvement, under different values of the regularization parameter $\eta$. \Cref{fig:training_stability} presents the results of five training runs on the expert dataset of the navigation task, each with a different value of $\eta$. The results show that the regularization term plays a crucial role in stabilizing the training dynamics. Lower values of $\eta$ lead to oscillations in policy performance and, in some cases, even result in divergence, as seen in the unregularized case. In contrast, higher values of $\eta$ reduce performance fluctuations between iterations, contributing to more stable and consistent learning progress.

\section{Conclusion}
We present Offline Munchausen Mirror Descent (\algname), a novel algorithm designed for learning equilibrium policies in mean-field games using only offline data. This approach addresses the limitations of existing methods that rely on costly and often impractical online interactions. By leveraging importance sampling and $Q$-value regularization techniques, \algname provides an efficient means to approximate the mean-field distribution from static datasets, ensuring scalability and robustness in complex environments. Our empirical evaluations demonstrated the algorithm's strong performance across two common benchmarks for MFGs, even in scenarios with limited data coverage or sub-optimal datasets. With its ability to scale and adapt to real-world multi-agent systems, Off-MMD opens new avenues for applying RL based algorithms for MFGs to settings where online experimentation is infeasible, irresponsible or difficult to model. We believe that this work lays the foundation for future research into offline learning methods for complex, large-scale multi-agent interactions, bridging the gap between offline RL and MFGs.
Future research directions include applications to real-world use-cases such as location recommendation systems or traffic routing, two problem domains suffering from overcrowding effects due to selfish agents.

\subsection{Limitations}

While \algname marks a first step towards scalable offline RL algorithms for MFGs, it is currently limited to environments where the dynamics are independent of the mean-field. This assumption restricts its applicability to scenarios where the environment dynamics are not arbitrarily influenced by the collective behavior of agents. Addressing this limitation offers a promising avenue for future research. One potential solution could involve adapting model-based algorithms specifically designed for offline RL settings, to handle mean-field dependencies in dynamics. Such advancements would broaden the scope of \algname, making it applicable to a wider range of multi-agent systems where interactions between agents and the environment are more intertwined.
%%%%%%%%%%%%%%%%%%%%%%%%%%%%%%%%%%%%%%%%%%%%%%%%%%%%%%%%%%%%%%%%%%%%%%%%

%%% The acknowledgments section is defined using the "acks" environment
%%% (rather than an unnumbered section). The use of this environment 
%%% ensures the proper identification of the section in the article 
%%% metadata as well as the consistent spelling of the heading.

%%%%%%%%%%%%%%%%%%%%%%%%%%%%%%%%%%%%%%%%%%%%%%%%%%%%%%%%%%%%%%%%%%%%%%%%

%%% The next two lines define, first, the bibliography style to be 
%%% applied, and, second, the bibliography file to be used.

\bibliographystyle{ACM-Reference-Format} 
\bibliography{paper}


\begin{thebibliography}{42}

%%% the Web address.

\ifx \showCODEN    \undefined \def \showCODEN     #1{\unskip}     \fi
\ifx \showDOI      \undefined \def \showDOI       #1{#1}\fi
\ifx \showISBNx    \undefined \def \showISBNx     #1{\unskip}     \fi
\ifx \showISBNxiii \undefined \def \showISBNxiii  #1{\unskip}     \fi
\ifx \showISSN     \undefined \def \showISSN      #1{\unskip}     \fi
\ifx \showLCCN     \undefined \def \showLCCN      #1{\unskip}     \fi
\ifx \shownote     \undefined \def \shownote      #1{#1}          \fi
\ifx \showarticletitle \undefined \def \showarticletitle #1{#1}   \fi
\ifx \showURL      \undefined \def \showURL       {\relax}        \fi
% invisible to TeX
\providecommand\bibfield[2]{#2}
\providecommand\bibinfo[2]{#2}
\providecommand\natexlab[1]{#1}
\providecommand\showeprint[2][]{arXiv:#2}

\bibitem[\protect\citeauthoryear{Achdou, Camilli, and Capuzzo-Dolcetta}{Achdou et~al\mbox{.}}{2012}]%
        {achdou2012mfg_numerical}
\bibfield{author}{\bibinfo{person}{Yves Achdou}, \bibinfo{person}{Fabio Camilli}, {and} \bibinfo{person}{Italo Capuzzo-Dolcetta}.} \bibinfo{year}{2012}\natexlab{}.
\newblock \showarticletitle{Mean Field Games: Numerical Methods for the Planning Problem}.
\newblock \bibinfo{journal}{\emph{SIAM Journal on Control and Optimization}} \bibinfo{volume}{50}, \bibinfo{number}{1} (\bibinfo{year}{2012}), \bibinfo{pages}{77--109}.
\newblock
\urldef\tempurl%
\url{https://doi.org/10.1137/100790069}
\showDOI{\tempurl}
\showeprint{https://doi.org/10.1137/100790069}


\bibitem[\protect\citeauthoryear{Achdou and Capuzzo-Dolcetta}{Achdou and Capuzzo-Dolcetta}{2010}]%
        {achdou2010mfg_numerical}
\bibfield{author}{\bibinfo{person}{Yves Achdou} {and} \bibinfo{person}{Italo Capuzzo-Dolcetta}.} \bibinfo{year}{2010}\natexlab{}.
\newblock \showarticletitle{Mean Field Games: Numerical Methods}.
\newblock \bibinfo{journal}{\emph{SIAM J. Numer. Anal.}} \bibinfo{volume}{48}, \bibinfo{number}{3} (\bibinfo{year}{2010}), \bibinfo{pages}{1136--1162}.
\newblock
\urldef\tempurl%
\url{https://doi.org/10.1137/090758477}
\showDOI{\tempurl}
\showeprint{https://doi.org/10.1137/090758477}


\bibitem[\protect\citeauthoryear{Bradbury, Frostig, Hawkins, Johnson, Leary, Maclaurin, Necula, Paszke, Vander{P}las, Wanderman-{M}ilne, and Zhang}{Bradbury et~al\mbox{.}}{2018}]%
        {jax2018github}
\bibfield{author}{\bibinfo{person}{James Bradbury}, \bibinfo{person}{Roy Frostig}, \bibinfo{person}{Peter Hawkins}, \bibinfo{person}{Matthew~James Johnson}, \bibinfo{person}{Chris Leary}, \bibinfo{person}{Dougal Maclaurin}, \bibinfo{person}{George Necula}, \bibinfo{person}{Adam Paszke}, \bibinfo{person}{Jake Vander{P}las}, \bibinfo{person}{Skye Wanderman-{M}ilne}, {and} \bibinfo{person}{Qiao Zhang}.} \bibinfo{year}{2018}\natexlab{}.
\newblock \bibinfo{booktitle}{\emph{{JAX}: composable transformations of {P}ython+{N}um{P}y programs}}.
\newblock
\urldef\tempurl%
\url{http://github.com/jax-ml/jax}
\showURL{%
\tempurl}


\bibitem[\protect\citeauthoryear{Brown}{Brown}{1951}]%
        {brown51fictitious}
\bibfield{author}{\bibinfo{person}{George~W. Brown}.} \bibinfo{year}{1951}\natexlab{}.
\newblock \showarticletitle{Iterative Solution of Games by Fictitious Play}.
\newblock In \bibinfo{booktitle}{\emph{Activity Analysis of Production and Allocation}}, \bibfield{editor}{\bibinfo{person}{T.~C. Koopmans}} (Ed.). \bibinfo{publisher}{Wiley}, \bibinfo{address}{New York}.
\newblock


\bibitem[\protect\citeauthoryear{{Cacace, Simone}, {Camilli, Fabio}, and {Goffi, Alessandro}}{{Cacace, Simone} et~al\mbox{.}}{2021}]%
        {cacace2021policy_iteration}
\bibfield{author}{\bibinfo{person}{{Cacace, Simone}}, \bibinfo{person}{{Camilli, Fabio}}, {and} \bibinfo{person}{{Goffi, Alessandro}}.} \bibinfo{year}{2021}\natexlab{}.
\newblock \showarticletitle{A policy iteration method for mean field games}.
\newblock \bibinfo{journal}{\emph{ESAIM: COCV}}  \bibinfo{volume}{27} (\bibinfo{year}{2021}), \bibinfo{pages}{85}.
\newblock
\urldef\tempurl%
\url{https://doi.org/10.1051/cocv/2021081}
\showDOI{\tempurl}


\bibitem[\protect\citeauthoryear{Cardaliaguet and Hadikhanloo}{Cardaliaguet and Hadikhanloo}{2017}]%
        {cardaliaguet2017fictitious}
\bibfield{author}{\bibinfo{person}{Pierre Cardaliaguet} {and} \bibinfo{person}{Saeed Hadikhanloo}.} \bibinfo{year}{2017}\natexlab{}.
\newblock \showarticletitle{Learning in mean field games: {The} fictitious play}.
\newblock \bibinfo{journal}{\emph{ESAIM: Control, Optimisation and Calculus of Variations}} \bibinfo{volume}{23}, \bibinfo{number}{2} (\bibinfo{year}{2017}), \bibinfo{pages}{569--591}.
\newblock
\urldef\tempurl%
\url{https://doi.org/10.1051/cocv/2016004}
\showDOI{\tempurl}


\bibitem[\protect\citeauthoryear{Carmona, Cooney, Graves, and Lauriere}{Carmona et~al\mbox{.}}{2019}]%
        {carmona2019stochasticgraphongamesi}
\bibfield{author}{\bibinfo{person}{Rene Carmona}, \bibinfo{person}{Daniel Cooney}, \bibinfo{person}{Christy Graves}, {and} \bibinfo{person}{Mathieu Lauriere}.} \bibinfo{year}{2019}\natexlab{}.
\newblock \bibinfo{title}{Stochastic Graphon Games: I. The Static Case}.
\newblock
\newblock
\showeprint[arxiv]{1911.10664}~[math.OC]
\urldef\tempurl%
\url{https://arxiv.org/abs/1911.10664}
\showURL{%
\tempurl}


\bibitem[\protect\citeauthoryear{Carmona, Laurière, and Tan}{Carmona et~al\mbox{.}}{2021}]%
        {carmona2021modelfreemeanfieldreinforcementlearning}
\bibfield{author}{\bibinfo{person}{René Carmona}, \bibinfo{person}{Mathieu Laurière}, {and} \bibinfo{person}{Zongjun Tan}.} \bibinfo{year}{2021}\natexlab{}.
\newblock \bibinfo{title}{Model-Free Mean-Field Reinforcement Learning: Mean-Field MDP and Mean-Field Q-Learning}.
\newblock
\newblock
\showeprint[arxiv]{1910.12802}~[math.OC]
\urldef\tempurl%
\url{https://arxiv.org/abs/1910.12802}
\showURL{%
\tempurl}


\bibitem[\protect\citeauthoryear{Chen, Li, Wang, Yang, Wang, and Zhao}{Chen et~al\mbox{.}}{2021}]%
        {chen2021pessimism}
\bibfield{author}{\bibinfo{person}{Minshuo Chen}, \bibinfo{person}{Yan Li}, \bibinfo{person}{Ethan Wang}, \bibinfo{person}{Zhuoran Yang}, \bibinfo{person}{Zhaoran Wang}, {and} \bibinfo{person}{Tuo Zhao}.} \bibinfo{year}{2021}\natexlab{}.
\newblock \showarticletitle{Pessimism Meets Invariance: Provably Efficient Offline Mean-Field Multi-Agent RL}. In \bibinfo{booktitle}{\emph{Advances in Neural Information Processing Systems}}, \bibfield{editor}{\bibinfo{person}{M.~Ranzato}, \bibinfo{person}{A.~Beygelzimer}, \bibinfo{person}{Y.~Dauphin}, \bibinfo{person}{P.S. Liang}, {and} \bibinfo{person}{J.~Wortman Vaughan}} (Eds.), Vol.~\bibinfo{volume}{34}. \bibinfo{publisher}{Curran Associates, Inc.}, \bibinfo{pages}{17913--17926}.
\newblock
\urldef\tempurl%
\url{https://proceedings.neurips.cc/paper_files/paper/2021/file/9559fc73b13fa721a816958488a5b449-Paper.pdf}
\showURL{%
\tempurl}


\bibitem[\protect\citeauthoryear{Chen, Zhang, Liu, and Hu}{Chen et~al\mbox{.}}{2022}]%
        {chen2022mfg-irl}
\bibfield{author}{\bibinfo{person}{Yang Chen}, \bibinfo{person}{Libo Zhang}, \bibinfo{person}{Jiamou Liu}, {and} \bibinfo{person}{Shuyue Hu}.} \bibinfo{year}{2022}\natexlab{}.
\newblock \showarticletitle{Individual-Level Inverse Reinforcement Learning for Mean Field Games}. In \bibinfo{booktitle}{\emph{Proceedings of the 21st International Conference on Autonomous Agents and Multiagent Systems}} (Virtual Event, New Zealand) \emph{(\bibinfo{series}{AAMAS '22})}. \bibinfo{publisher}{International Foundation for Autonomous Agents and Multiagent Systems}, \bibinfo{address}{Richland, SC}, \bibinfo{pages}{253–262}.
\newblock
\showISBNx{9781450392136}


\bibitem[\protect\citeauthoryear{Chen, Zhang, Liu, and Witbrock}{Chen et~al\mbox{.}}{2023}]%
        {chen2023mfg-irl}
\bibfield{author}{\bibinfo{person}{Yang Chen}, \bibinfo{person}{Libo Zhang}, \bibinfo{person}{Jiamou Liu}, {and} \bibinfo{person}{Michael Witbrock}.} \bibinfo{year}{2023}\natexlab{}.
\newblock \showarticletitle{Adversarial Inverse Reinforcement Learning for Mean Field Games}. In \bibinfo{booktitle}{\emph{Proceedings of the 2023 International Conference on Autonomous Agents and Multiagent Systems}} (London, United Kingdom) \emph{(\bibinfo{series}{AAMAS '23})}. \bibinfo{publisher}{International Foundation for Autonomous Agents and Multiagent Systems}, \bibinfo{address}{Richland, SC}, \bibinfo{pages}{1088–1096}.
\newblock
\showISBNx{9781450394321}


\bibitem[\protect\citeauthoryear{Cui and Koeppl}{Cui and Koeppl}{2021}]%
        {cui2021approximately_solving}
\bibfield{author}{\bibinfo{person}{Kai Cui} {and} \bibinfo{person}{Heinz Koeppl}.} \bibinfo{year}{2021}\natexlab{}.
\newblock \showarticletitle{Approximately Solving Mean Field Games via Entropy-Regularized Deep Reinforcement Learning}. In \bibinfo{booktitle}{\emph{Proceedings of The 24th International Conference on Artificial Intelligence and Statistics}} \emph{(\bibinfo{series}{Proceedings of Machine Learning Research}, Vol.~\bibinfo{volume}{130})}, \bibfield{editor}{\bibinfo{person}{Arindam Banerjee} {and} \bibinfo{person}{Kenji Fukumizu}} (Eds.). \bibinfo{publisher}{PMLR}, \bibinfo{pages}{1909--1917}.
\newblock
\urldef\tempurl%
\url{https://proceedings.mlr.press/v130/cui21a.html}
\showURL{%
\tempurl}


\bibitem[\protect\citeauthoryear{Fabian, Cui, and Koeppl}{Fabian et~al\mbox{.}}{2023}]%
        {fabian2023graphon}
\bibfield{author}{\bibinfo{person}{Christian Fabian}, \bibinfo{person}{Kai Cui}, {and} \bibinfo{person}{Heinz Koeppl}.} \bibinfo{year}{2023}\natexlab{}.
\newblock \showarticletitle{Learning Sparse Graphon Mean Field Games}. In \bibinfo{booktitle}{\emph{Proceedings of The 26th International Conference on Artificial Intelligence and Statistics}} \emph{(\bibinfo{series}{Proceedings of Machine Learning Research}, Vol.~\bibinfo{volume}{206})}, \bibfield{editor}{\bibinfo{person}{Francisco Ruiz}, \bibinfo{person}{Jennifer Dy}, {and} \bibinfo{person}{Jan-Willem van~de Meent}} (Eds.). \bibinfo{publisher}{PMLR}, \bibinfo{pages}{4486--4514}.
\newblock
\urldef\tempurl%
\url{https://proceedings.mlr.press/v206/fabian23a.html}
\showURL{%
\tempurl}


\bibitem[\protect\citeauthoryear{Guo, Hu, Xu, and Zhang}{Guo et~al\mbox{.}}{2019}]%
        {guo2019learning}
\bibfield{author}{\bibinfo{person}{Xin Guo}, \bibinfo{person}{Anran Hu}, \bibinfo{person}{Renyuan Xu}, {and} \bibinfo{person}{Junzi Zhang}.} \bibinfo{year}{2019}\natexlab{}.
\newblock \showarticletitle{Learning Mean-Field Games}. In \bibinfo{booktitle}{\emph{Advances in Neural Information Processing Systems}}, \bibfield{editor}{\bibinfo{person}{H.~Wallach}, \bibinfo{person}{H.~Larochelle}, \bibinfo{person}{A.~Beygelzimer}, \bibinfo{person}{F.~d\textquotesingle Alch\'{e}-Buc}, \bibinfo{person}{E.~Fox}, {and} \bibinfo{person}{R.~Garnett}} (Eds.), Vol.~\bibinfo{volume}{32}. \bibinfo{publisher}{Curran Associates, Inc.}
\newblock
\urldef\tempurl%
\url{https://proceedings.neurips.cc/paper_files/paper/2019/file/030e65da2b1c944090548d36b244b28d-Paper.pdf}
\showURL{%
\tempurl}


\bibitem[\protect\citeauthoryear{Heinrich, Lanctot, and Silver}{Heinrich et~al\mbox{.}}{2015}]%
        {pmlr-v37-heinrich15}
\bibfield{author}{\bibinfo{person}{Johannes Heinrich}, \bibinfo{person}{Marc Lanctot}, {and} \bibinfo{person}{David Silver}.} \bibinfo{year}{2015}\natexlab{}.
\newblock \showarticletitle{Fictitious Self-Play in Extensive-Form Games}. In \bibinfo{booktitle}{\emph{Proceedings of the 32nd International Conference on Machine Learning}} \emph{(\bibinfo{series}{Proceedings of Machine Learning Research}, Vol.~\bibinfo{volume}{37})}, \bibfield{editor}{\bibinfo{person}{Francis Bach} {and} \bibinfo{person}{David Blei}} (Eds.). \bibinfo{publisher}{PMLR}, \bibinfo{address}{Lille, France}, \bibinfo{pages}{805--813}.
\newblock
\urldef\tempurl%
\url{https://proceedings.mlr.press/v37/heinrich15.html}
\showURL{%
\tempurl}


\bibitem[\protect\citeauthoryear{Huang, Yardim, and He}{Huang et~al\mbox{.}}{2024}]%
        {huang2024statistical}
\bibfield{author}{\bibinfo{person}{Jiawei Huang}, \bibinfo{person}{Batuhan Yardim}, {and} \bibinfo{person}{Niao He}.} \bibinfo{year}{2024}\natexlab{}.
\newblock \showarticletitle{On the Statistical Efficiency of Mean-Field Reinforcement Learning with General Function Approximation}. In \bibinfo{booktitle}{\emph{Proceedings of The 27th International Conference on Artificial Intelligence and Statistics}} \emph{(\bibinfo{series}{Proceedings of Machine Learning Research}, Vol.~\bibinfo{volume}{238})}, \bibfield{editor}{\bibinfo{person}{Sanjoy Dasgupta}, \bibinfo{person}{Stephan Mandt}, {and} \bibinfo{person}{Yingzhen Li}} (Eds.). \bibinfo{publisher}{PMLR}, \bibinfo{pages}{289--297}.
\newblock
\urldef\tempurl%
\url{https://proceedings.mlr.press/v238/huang24a.html}
\showURL{%
\tempurl}


\bibitem[\protect\citeauthoryear{Huang, Malhame, and Caines}{Huang et~al\mbox{.}}{2006}]%
        {hang2006large}
\bibfield{author}{\bibinfo{person}{Minyi Huang}, \bibinfo{person}{Roland Malhame}, {and} \bibinfo{person}{Peter Caines}.} \bibinfo{year}{2006}\natexlab{}.
\newblock \showarticletitle{Large population stochastic dynamic games: Closed-loop McKean-Vlasov systems and the Nash certainty equivalence principle}.
\newblock \bibinfo{journal}{\emph{Commun. Inf. Syst.}}  \bibinfo{volume}{6} (\bibinfo{date}{01} \bibinfo{year}{2006}).
\newblock
\urldef\tempurl%
\url{https://doi.org/10.4310/CIS.2006.v6.n3.a5}
\showDOI{\tempurl}


\bibitem[\protect\citeauthoryear{Jusup, P\'{a}sztor, Janik, Zhang, Corman, Krause, and Bogunovic}{Jusup et~al\mbox{.}}{2024}]%
        {matej2024safe}
\bibfield{author}{\bibinfo{person}{Matej Jusup}, \bibinfo{person}{Barna P\'{a}sztor}, \bibinfo{person}{Tadeusz Janik}, \bibinfo{person}{Kenan Zhang}, \bibinfo{person}{Francesco Corman}, \bibinfo{person}{Andreas Krause}, {and} \bibinfo{person}{Ilija Bogunovic}.} \bibinfo{year}{2024}\natexlab{}.
\newblock \showarticletitle{Safe Model-Based Multi-Agent Mean-Field Reinforcement Learning}. In \bibinfo{booktitle}{\emph{Proceedings of the 23rd International Conference on Autonomous Agents and Multiagent Systems}} (Auckland, New Zealand) \emph{(\bibinfo{series}{AAMAS '24})}. \bibinfo{publisher}{International Foundation for Autonomous Agents and Multiagent Systems}, \bibinfo{address}{Richland, SC}, \bibinfo{pages}{973–982}.
\newblock
\showISBNx{9798400704864}


\bibitem[\protect\citeauthoryear{Kidambi, Rajeswaran, Netrapalli, and Joachims}{Kidambi et~al\mbox{.}}{2020}]%
        {kidambi2020morel}
\bibfield{author}{\bibinfo{person}{Rahul Kidambi}, \bibinfo{person}{Aravind Rajeswaran}, \bibinfo{person}{Praneeth Netrapalli}, {and} \bibinfo{person}{Thorsten Joachims}.} \bibinfo{year}{2020}\natexlab{}.
\newblock \showarticletitle{MOReL: Model-Based Offline Reinforcement Learning}. In \bibinfo{booktitle}{\emph{Advances in Neural Information Processing Systems}}, \bibfield{editor}{\bibinfo{person}{H.~Larochelle}, \bibinfo{person}{M.~Ranzato}, \bibinfo{person}{R.~Hadsell}, \bibinfo{person}{M.F. Balcan}, {and} \bibinfo{person}{H.~Lin}} (Eds.), Vol.~\bibinfo{volume}{33}. \bibinfo{publisher}{Curran Associates, Inc.}, \bibinfo{pages}{21810--21823}.
\newblock
\urldef\tempurl%
\url{https://proceedings.neurips.cc/paper_files/paper/2020/file/f7efa4f864ae9b88d43527f4b14f750f-Paper.pdf}
\showURL{%
\tempurl}


\bibitem[\protect\citeauthoryear{Kostrikov}{Kostrikov}{2021}]%
        {jaxrl}
\bibfield{author}{\bibinfo{person}{Ilya Kostrikov}.} \bibinfo{year}{2021}\natexlab{}.
\newblock \bibinfo{title}{{JAXRL: Implementations of Reinforcement Learning algorithms in JAX}}.
\newblock
\newblock
\urldef\tempurl%
\url{https://doi.org/10.5281/zenodo.5535154}
\showDOI{\tempurl}


\bibitem[\protect\citeauthoryear{Kumar, Zhou, Tucker, and Levine}{Kumar et~al\mbox{.}}{2020}]%
        {kumar2020cql}
\bibfield{author}{\bibinfo{person}{Aviral Kumar}, \bibinfo{person}{Aurick Zhou}, \bibinfo{person}{George Tucker}, {and} \bibinfo{person}{Sergey Levine}.} \bibinfo{year}{2020}\natexlab{}.
\newblock \showarticletitle{Conservative Q-Learning for Offline Reinforcement Learning}. In \bibinfo{booktitle}{\emph{Advances in Neural Information Processing Systems}}, \bibfield{editor}{\bibinfo{person}{H.~Larochelle}, \bibinfo{person}{M.~Ranzato}, \bibinfo{person}{R.~Hadsell}, \bibinfo{person}{M.F. Balcan}, {and} \bibinfo{person}{H.~Lin}} (Eds.), Vol.~\bibinfo{volume}{33}. \bibinfo{publisher}{Curran Associates, Inc.}, \bibinfo{pages}{1179--1191}.
\newblock
\urldef\tempurl%
\url{https://proceedings.neurips.cc/paper_files/paper/2020/file/0d2b2061826a5df3221116a5085a6052-Paper.pdf}
\showURL{%
\tempurl}


\bibitem[\protect\citeauthoryear{Lanctot, Lockhart, Lespiau, Zambaldi, Upadhyay, P\'{e}rolat, Srinivasan, Timbers, Tuyls, Omidshafiei, Hennes, Morrill, Muller, Ewalds, Faulkner, Kram\'{a}r, Vylder, Saeta, Bradbury, Ding, Borgeaud, Lai, Schrittwieser, Anthony, Hughes, Danihelka, and Ryan-Davis}{Lanctot et~al\mbox{.}}{2019}]%
        {lanctot2019openspiel}
\bibfield{author}{\bibinfo{person}{Marc Lanctot}, \bibinfo{person}{Edward Lockhart}, \bibinfo{person}{Jean-Baptiste Lespiau}, \bibinfo{person}{Vinicius Zambaldi}, \bibinfo{person}{Satyaki Upadhyay}, \bibinfo{person}{Julien P\'{e}rolat}, \bibinfo{person}{Sriram Srinivasan}, \bibinfo{person}{Finbarr Timbers}, \bibinfo{person}{Karl Tuyls}, \bibinfo{person}{Shayegan Omidshafiei}, \bibinfo{person}{Daniel Hennes}, \bibinfo{person}{Dustin Morrill}, \bibinfo{person}{Paul Muller}, \bibinfo{person}{Timo Ewalds}, \bibinfo{person}{Ryan Faulkner}, \bibinfo{person}{J\'{a}nos Kram\'{a}r}, \bibinfo{person}{Bart~De Vylder}, \bibinfo{person}{Brennan Saeta}, \bibinfo{person}{James Bradbury}, \bibinfo{person}{David Ding}, \bibinfo{person}{Sebastian Borgeaud}, \bibinfo{person}{Matthew Lai}, \bibinfo{person}{Julian Schrittwieser}, \bibinfo{person}{Thomas Anthony}, \bibinfo{person}{Edward Hughes}, \bibinfo{person}{Ivo Danihelka}, {and} \bibinfo{person}{Jonah Ryan-Davis}.} \bibinfo{year}{2019}\natexlab{}.
\newblock \showarticletitle{{OpenSpiel}: A Framework for Reinforcement Learning in Games}.
\newblock \bibinfo{journal}{\emph{CoRR}}  \bibinfo{volume}{abs/1908.09453} (\bibinfo{year}{2019}).
\newblock
\showeprint[arxiv]{1908.09453}~[cs.LG]
\urldef\tempurl%
\url{http://arxiv.org/abs/1908.09453}
\showURL{%
\tempurl}


\bibitem[\protect\citeauthoryear{Lasry and Lions}{Lasry and Lions}{2007}]%
        {lasry2007mfg}
\bibfield{author}{\bibinfo{person}{J.~M. Lasry} {and} \bibinfo{person}{Pierre-Louis Lions}.} \bibinfo{year}{2007}\natexlab{}.
\newblock \showarticletitle{Mean field games}.
\newblock \bibinfo{journal}{\emph{Japanese Journal of Mathematics}}  \bibinfo{volume}{2} (\bibinfo{year}{2007}), \bibinfo{pages}{229--260}.
\newblock
\urldef\tempurl%
\url{https://api.semanticscholar.org/CorpusID:1963678}
\showURL{%
\tempurl}


\bibitem[\protect\citeauthoryear{Lauri{\`e}re, Perrin, Girgin, Muller, Jain, Cabannes, Piliouras, P{\'e}rolat, Elie, Pietquin, et~al\mbox{.}}{Lauri{\`e}re et~al\mbox{.}}{2022}]%
        {lauriere2022scalable}
\bibfield{author}{\bibinfo{person}{Mathieu Lauri{\`e}re}, \bibinfo{person}{Sarah Perrin}, \bibinfo{person}{Sertan Girgin}, \bibinfo{person}{Paul Muller}, \bibinfo{person}{Ayush Jain}, \bibinfo{person}{Theophile Cabannes}, \bibinfo{person}{Georgios Piliouras}, \bibinfo{person}{Julien P{\'e}rolat}, \bibinfo{person}{Romuald Elie}, \bibinfo{person}{Olivier Pietquin}, {et~al\mbox{.}}} \bibinfo{year}{2022}\natexlab{}.
\newblock \showarticletitle{Scalable deep reinforcement learning algorithms for mean field games}. In \bibinfo{booktitle}{\emph{International Conference on Machine Learning}}. PMLR, \bibinfo{pages}{12078--12095}.
\newblock


\bibitem[\protect\citeauthoryear{Laurière, Perrin, Pérolat, Girgin, Muller, Élie, Geist, and Pietquin}{Laurière et~al\mbox{.}}{2024}]%
        {lauriere2024learningmeanfieldgames}
\bibfield{author}{\bibinfo{person}{Mathieu Laurière}, \bibinfo{person}{Sarah Perrin}, \bibinfo{person}{Julien Pérolat}, \bibinfo{person}{Sertan Girgin}, \bibinfo{person}{Paul Muller}, \bibinfo{person}{Romuald Élie}, \bibinfo{person}{Matthieu Geist}, {and} \bibinfo{person}{Olivier Pietquin}.} \bibinfo{year}{2024}\natexlab{}.
\newblock \bibinfo{title}{Learning in Mean Field Games: A Survey}.
\newblock
\newblock
\showeprint[arxiv]{2205.12944}~[cs.LG]
\urldef\tempurl%
\url{https://arxiv.org/abs/2205.12944}
\showURL{%
\tempurl}


\bibitem[\protect\citeauthoryear{Lee, Jeon, Lee, Pineau, and Kim}{Lee et~al\mbox{.}}{2021}]%
        {lee2021optidice}
\bibfield{author}{\bibinfo{person}{Jongmin Lee}, \bibinfo{person}{Wonseok Jeon}, \bibinfo{person}{Byungjun Lee}, \bibinfo{person}{Joelle Pineau}, {and} \bibinfo{person}{Kee-Eung Kim}.} \bibinfo{year}{2021}\natexlab{}.
\newblock \showarticletitle{OptiDICE: Offline Policy Optimization via Stationary Distribution Correction Estimation}. In \bibinfo{booktitle}{\emph{Proceedings of the 38th International Conference on Machine Learning}} \emph{(\bibinfo{series}{Proceedings of Machine Learning Research}, Vol.~\bibinfo{volume}{139})}, \bibfield{editor}{\bibinfo{person}{Marina Meila} {and} \bibinfo{person}{Tong Zhang}} (Eds.). \bibinfo{publisher}{PMLR}, \bibinfo{pages}{6120--6130}.
\newblock
\urldef\tempurl%
\url{https://proceedings.mlr.press/v139/lee21f.html}
\showURL{%
\tempurl}


\bibitem[\protect\citeauthoryear{Mnih, Kavukcuoglu, Silver, Rusu, Veness, Bellemare, Graves, Riedmiller, Fidjeland, Ostrovski, Petersen, Beattie, Sadik, Antonoglou, King, Kumaran, Wierstra, Legg, and Hassabis}{Mnih et~al\mbox{.}}{2015}]%
        {mnih2015dqn}
\bibfield{author}{\bibinfo{person}{Volodymyr Mnih}, \bibinfo{person}{Koray Kavukcuoglu}, \bibinfo{person}{David Silver}, \bibinfo{person}{Andrei~A. Rusu}, \bibinfo{person}{Joel Veness}, \bibinfo{person}{Marc~G. Bellemare}, \bibinfo{person}{Alex Graves}, \bibinfo{person}{Martin~A. Riedmiller}, \bibinfo{person}{Andreas~Kirkeby Fidjeland}, \bibinfo{person}{Georg Ostrovski}, \bibinfo{person}{Stig Petersen}, \bibinfo{person}{Charlie Beattie}, \bibinfo{person}{Amir Sadik}, \bibinfo{person}{Ioannis Antonoglou}, \bibinfo{person}{Helen King}, \bibinfo{person}{Dharshan Kumaran}, \bibinfo{person}{Daan Wierstra}, \bibinfo{person}{Shane Legg}, {and} \bibinfo{person}{Demis Hassabis}.} \bibinfo{year}{2015}\natexlab{}.
\newblock \showarticletitle{Human-level control through deep reinforcement learning}.
\newblock \bibinfo{journal}{\emph{Nature}}  \bibinfo{volume}{518} (\bibinfo{year}{2015}), \bibinfo{pages}{529--533}.
\newblock
\urldef\tempurl%
\url{https://api.semanticscholar.org/CorpusID:205242740}
\showURL{%
\tempurl}


\bibitem[\protect\citeauthoryear{Nachum, Chow, Dai, and Li}{Nachum et~al\mbox{.}}{2019}]%
        {nachum2020dualdice}
\bibfield{author}{\bibinfo{person}{Ofir Nachum}, \bibinfo{person}{Yinlam Chow}, \bibinfo{person}{Bo Dai}, {and} \bibinfo{person}{Lihong Li}.} \bibinfo{year}{2019}\natexlab{}.
\newblock \showarticletitle{DualDICE: Behavior-Agnostic Estimation of Discounted Stationary Distribution Corrections}. In \bibinfo{booktitle}{\emph{Advances in Neural Information Processing Systems}}, \bibfield{editor}{\bibinfo{person}{H.~Wallach}, \bibinfo{person}{H.~Larochelle}, \bibinfo{person}{A.~Beygelzimer}, \bibinfo{person}{F.~d\textquotesingle Alch\'{e}-Buc}, \bibinfo{person}{E.~Fox}, {and} \bibinfo{person}{R.~Garnett}} (Eds.), Vol.~\bibinfo{volume}{32}. \bibinfo{publisher}{Curran Associates, Inc.}
\newblock
\urldef\tempurl%
\url{https://proceedings.neurips.cc/paper_files/paper/2019/file/cf9a242b70f45317ffd281241fa66502-Paper.pdf}
\showURL{%
\tempurl}


\bibitem[\protect\citeauthoryear{P{\'a}sztor, Krause, and Bogunovic}{P{\'a}sztor et~al\mbox{.}}{2023}]%
        {pasztor2023efficient}
\bibfield{author}{\bibinfo{person}{Barna P{\'a}sztor}, \bibinfo{person}{Andreas Krause}, {and} \bibinfo{person}{Ilija Bogunovic}.} \bibinfo{year}{2023}\natexlab{}.
\newblock \showarticletitle{Efficient Model-Based Multi-Agent Mean-Field Reinforcement Learning}.
\newblock \bibinfo{journal}{\emph{Transactions on Machine Learning Research}} (\bibinfo{year}{2023}).
\newblock
\showISSN{2835-8856}
\urldef\tempurl%
\url{https://openreview.net/forum?id=gvcDSDYUZx}
\showURL{%
\tempurl}


\bibitem[\protect\citeauthoryear{P\'{e}rolat, Perrin, Elie, Lauri\`{e}re, Piliouras, Geist, Tuyls, and Pietquin}{P\'{e}rolat et~al\mbox{.}}{2022}]%
        {perolat2022omd}
\bibfield{author}{\bibinfo{person}{Julien P\'{e}rolat}, \bibinfo{person}{Sarah Perrin}, \bibinfo{person}{Romuald Elie}, \bibinfo{person}{Mathieu Lauri\`{e}re}, \bibinfo{person}{Georgios Piliouras}, \bibinfo{person}{Matthieu Geist}, \bibinfo{person}{Karl Tuyls}, {and} \bibinfo{person}{Olivier Pietquin}.} \bibinfo{year}{2022}\natexlab{}.
\newblock \showarticletitle{Scaling Mean Field Games by Online Mirror Descent}. In \bibinfo{booktitle}{\emph{Proceedings of the 21st International Conference on Autonomous Agents and Multiagent Systems}} (Virtual Event, New Zealand) \emph{(\bibinfo{series}{AAMAS '22})}. \bibinfo{publisher}{International Foundation for Autonomous Agents and Multiagent Systems}, \bibinfo{address}{Richland, SC}, \bibinfo{pages}{1028–1037}.
\newblock
\showISBNx{9781450392136}


\bibitem[\protect\citeauthoryear{Perrin}{Perrin}{2022}]%
        {perrin2022scaling}
\bibfield{author}{\bibinfo{person}{Sarah Perrin}.} \bibinfo{year}{2022}\natexlab{}.
\newblock \emph{\bibinfo{title}{{Scaling up Multi-agent Reinforcement Learning with Mean Field Games and Vice-versa}}}.
\newblock Theses. \bibinfo{school}{{Universit{\'e} de Lille}}.
\newblock
\urldef\tempurl%
\url{https://theses.hal.science/tel-04284876}
\showURL{%
\tempurl}


\bibitem[\protect\citeauthoryear{Perrin, Laurière, Pérolat, Geist, Élie, and Pietquin}{Perrin et~al\mbox{.}}{2021}]%
        {perrin2021flock}
\bibfield{author}{\bibinfo{person}{Sarah Perrin}, \bibinfo{person}{Mathieu Laurière}, \bibinfo{person}{Julien Pérolat}, \bibinfo{person}{Matthieu Geist}, \bibinfo{person}{Romuald Élie}, {and} \bibinfo{person}{Olivier Pietquin}.} \bibinfo{year}{2021}\natexlab{}.
\newblock \showarticletitle{Mean Field Games Flock! The Reinforcement Learning Way}. In \bibinfo{booktitle}{\emph{Proceedings of the Thirtieth International Joint Conference on Artificial Intelligence, {IJCAI-21}}}, \bibfield{editor}{\bibinfo{person}{Zhi-Hua Zhou}} (Ed.). \bibinfo{publisher}{International Joint Conferences on Artificial Intelligence Organization}, \bibinfo{pages}{356--362}.
\newblock
\urldef\tempurl%
\url{https://doi.org/10.24963/ijcai.2021/50}
\showDOI{\tempurl}
\newblock
\shownote{Main Track.}


\bibitem[\protect\citeauthoryear{Perrin, Perolat, Lauriere, Geist, Elie, and Pietquin}{Perrin et~al\mbox{.}}{2020}]%
        {perrin2020fictitious}
\bibfield{author}{\bibinfo{person}{Sarah Perrin}, \bibinfo{person}{Julien Perolat}, \bibinfo{person}{Mathieu Lauriere}, \bibinfo{person}{Matthieu Geist}, \bibinfo{person}{Romuald Elie}, {and} \bibinfo{person}{Olivier Pietquin}.} \bibinfo{year}{2020}\natexlab{}.
\newblock \showarticletitle{Fictitious Play for Mean Field Games: Continuous Time Analysis and Applications}. In \bibinfo{booktitle}{\emph{Advances in Neural Information Processing Systems}}, \bibfield{editor}{\bibinfo{person}{H.~Larochelle}, \bibinfo{person}{M.~Ranzato}, \bibinfo{person}{R.~Hadsell}, \bibinfo{person}{M.F. Balcan}, {and} \bibinfo{person}{H.~Lin}} (Eds.), Vol.~\bibinfo{volume}{33}. \bibinfo{publisher}{Curran Associates, Inc.}, \bibinfo{pages}{13199--13213}.
\newblock
\urldef\tempurl%
\url{https://proceedings.neurips.cc/paper_files/paper/2020/file/995ca733e3657ff9f5f3c823d73371e1-Paper.pdf}
\showURL{%
\tempurl}


\bibitem[\protect\citeauthoryear{Schweighofer, Dinu, Radler, Hofmarcher, Patil, Bitto-nemling, Eghbal-zadeh, and Hochreiter}{Schweighofer et~al\mbox{.}}{2022}]%
        {schweighofer2022dataser}
\bibfield{author}{\bibinfo{person}{Kajetan Schweighofer}, \bibinfo{person}{Marius-constantin Dinu}, \bibinfo{person}{Andreas Radler}, \bibinfo{person}{Markus Hofmarcher}, \bibinfo{person}{Vihang~Prakash Patil}, \bibinfo{person}{Angela Bitto-nemling}, \bibinfo{person}{Hamid Eghbal-zadeh}, {and} \bibinfo{person}{Sepp Hochreiter}.} \bibinfo{year}{2022}\natexlab{}.
\newblock \showarticletitle{A Dataset Perspective on Offline Reinforcement Learning}. In \bibinfo{booktitle}{\emph{Proceedings of The 1st Conference on Lifelong Learning Agents}} \emph{(\bibinfo{series}{Proceedings of Machine Learning Research}, Vol.~\bibinfo{volume}{199})}, \bibfield{editor}{\bibinfo{person}{Sarath Chandar}, \bibinfo{person}{Razvan Pascanu}, {and} \bibinfo{person}{Doina Precup}} (Eds.). \bibinfo{publisher}{PMLR}, \bibinfo{pages}{470--517}.
\newblock
\urldef\tempurl%
\url{https://proceedings.mlr.press/v199/schweighofer22a.html}
\showURL{%
\tempurl}


\bibitem[\protect\citeauthoryear{Subramanian, Taylor, Crowley, and Poupart}{Subramanian et~al\mbox{.}}{2022}]%
        {subramanian2022decentralized}
\bibfield{author}{\bibinfo{person}{Sriram~Ganapathi Subramanian}, \bibinfo{person}{Matthew~E. Taylor}, \bibinfo{person}{Mark Crowley}, {and} \bibinfo{person}{Pascal Poupart}.} \bibinfo{year}{2022}\natexlab{}.
\newblock \showarticletitle{Decentralized Mean Field Games}.
\newblock \bibinfo{journal}{\emph{Proceedings of the AAAI Conference on Artificial Intelligence}} \bibinfo{volume}{36}, \bibinfo{number}{9} (\bibinfo{date}{Jun.} \bibinfo{year}{2022}), \bibinfo{pages}{9439--9447}.
\newblock
\urldef\tempurl%
\url{https://doi.org/10.1609/aaai.v36i9.21176}
\showDOI{\tempurl}


\bibitem[\protect\citeauthoryear{Wen, Dai, Li, and Schuurmans}{Wen et~al\mbox{.}}{2020}]%
        {wen2020batch_stationary_distribution}
\bibfield{author}{\bibinfo{person}{Junfeng Wen}, \bibinfo{person}{Bo Dai}, \bibinfo{person}{Lihong Li}, {and} \bibinfo{person}{Dale Schuurmans}.} \bibinfo{year}{2020}\natexlab{}.
\newblock \showarticletitle{Batch stationary distribution estimation}. In \bibinfo{booktitle}{\emph{Proceedings of the 37th International Conference on Machine Learning}} \emph{(\bibinfo{series}{ICML'20})}. \bibinfo{publisher}{JMLR.org}, Article \bibinfo{articleno}{945}, \bibinfo{numpages}{11}~pages.
\newblock


\bibitem[\protect\citeauthoryear{Xie, Yang, Wang, and Minca}{Xie et~al\mbox{.}}{2021}]%
        {xie2021learning}
\bibfield{author}{\bibinfo{person}{Qiaomin Xie}, \bibinfo{person}{Zhuoran Yang}, \bibinfo{person}{Zhaoran Wang}, {and} \bibinfo{person}{Andreea Minca}.} \bibinfo{year}{2021}\natexlab{}.
\newblock \showarticletitle{Learning While Playing in Mean-Field Games: Convergence and Optimality}. In \bibinfo{booktitle}{\emph{Proceedings of the 38th International Conference on Machine Learning}} \emph{(\bibinfo{series}{Proceedings of Machine Learning Research}, Vol.~\bibinfo{volume}{139})}, \bibfield{editor}{\bibinfo{person}{Marina Meila} {and} \bibinfo{person}{Tong Zhang}} (Eds.). \bibinfo{publisher}{PMLR}, \bibinfo{pages}{11436--11447}.
\newblock
\urldef\tempurl%
\url{https://proceedings.mlr.press/v139/xie21g.html}
\showURL{%
\tempurl}


\bibitem[\protect\citeauthoryear{Xie, Ma, and Wang}{Xie et~al\mbox{.}}{2019}]%
        {xie2019mis}
\bibfield{author}{\bibinfo{person}{Tengyang Xie}, \bibinfo{person}{Yifei Ma}, {and} \bibinfo{person}{Yu-Xiang Wang}.} \bibinfo{year}{2019}\natexlab{}.
\newblock \showarticletitle{Towards Optimal Off-Policy Evaluation for Reinforcement Learning with Marginalized Importance Sampling}. In \bibinfo{booktitle}{\emph{Advances in Neural Information Processing Systems}}, Vol.~\bibinfo{volume}{32}.
\newblock
\urldef\tempurl%
\url{https://proceedings.neurips.cc/paper_files/paper/2019/file/4ffb0d2ba92f664c2281970110a2e071-Paper.pdf}
\showURL{%
\tempurl}


\bibitem[\protect\citeauthoryear{Yang, Ye, Trivedi, Xu, and Zha}{Yang et~al\mbox{.}}{2018b}]%
        {yang2018irl}
\bibfield{author}{\bibinfo{person}{Jiachen Yang}, \bibinfo{person}{Xiaojing Ye}, \bibinfo{person}{Rakshit Trivedi}, \bibinfo{person}{Huan Xu}, {and} \bibinfo{person}{Hongyuan Zha}.} \bibinfo{year}{2018}\natexlab{b}.
\newblock \showarticletitle{Deep Mean Field Games for Learning Optimal Behavior Policy of Large Populations}. In \bibinfo{booktitle}{\emph{International Conference on Learning Representations}}.
\newblock
\urldef\tempurl%
\url{https://openreview.net/forum?id=HktK4BeCZ}
\showURL{%
\tempurl}


\bibitem[\protect\citeauthoryear{Yang, Luo, Li, Zhou, Zhang, and Wang}{Yang et~al\mbox{.}}{2018a}]%
        {yang2018mf_marl}
\bibfield{author}{\bibinfo{person}{Yaodong Yang}, \bibinfo{person}{Rui Luo}, \bibinfo{person}{Minne Li}, \bibinfo{person}{Ming Zhou}, \bibinfo{person}{Weinan Zhang}, {and} \bibinfo{person}{Jun Wang}.} \bibinfo{year}{2018}\natexlab{a}.
\newblock \showarticletitle{Mean Field Multi-Agent Reinforcement Learning}. In \bibinfo{booktitle}{\emph{Proceedings of the 35th International Conference on Machine Learning}} \emph{(\bibinfo{series}{Proceedings of Machine Learning Research}, Vol.~\bibinfo{volume}{80})}, \bibfield{editor}{\bibinfo{person}{Jennifer Dy} {and} \bibinfo{person}{Andreas Krause}} (Eds.). \bibinfo{publisher}{PMLR}, \bibinfo{pages}{5571--5580}.
\newblock
\urldef\tempurl%
\url{https://proceedings.mlr.press/v80/yang18d.html}
\showURL{%
\tempurl}


\bibitem[\protect\citeauthoryear{Yu, Thomas, Yu, Ermon, Zou, Levine, Finn, and Ma}{Yu et~al\mbox{.}}{2020}]%
        {yu2020mopo}
\bibfield{author}{\bibinfo{person}{Tianhe Yu}, \bibinfo{person}{Garrett Thomas}, \bibinfo{person}{Lantao Yu}, \bibinfo{person}{Stefano Ermon}, \bibinfo{person}{James~Y Zou}, \bibinfo{person}{Sergey Levine}, \bibinfo{person}{Chelsea Finn}, {and} \bibinfo{person}{Tengyu Ma}.} \bibinfo{year}{2020}\natexlab{}.
\newblock \showarticletitle{MOPO: Model-based Offline Policy Optimization}. In \bibinfo{booktitle}{\emph{Advances in Neural Information Processing Systems}}, \bibfield{editor}{\bibinfo{person}{H.~Larochelle}, \bibinfo{person}{M.~Ranzato}, \bibinfo{person}{R.~Hadsell}, \bibinfo{person}{M.F. Balcan}, {and} \bibinfo{person}{H.~Lin}} (Eds.), Vol.~\bibinfo{volume}{33}. \bibinfo{publisher}{Curran Associates, Inc.}, \bibinfo{pages}{14129--14142}.
\newblock
\urldef\tempurl%
\url{https://proceedings.neurips.cc/paper_files/paper/2020/file/a322852ce0df73e204b7e67cbbef0d0a-Paper.pdf}
\showURL{%
\tempurl}


\bibitem[\protect\citeauthoryear{Zhang*, Dai*, Li, and Schuurmans}{Zhang* et~al\mbox{.}}{2020}]%
        {zhang2020GenDICE}
\bibfield{author}{\bibinfo{person}{Ruiyi Zhang*}, \bibinfo{person}{Bo Dai*}, \bibinfo{person}{Lihong Li}, {and} \bibinfo{person}{Dale Schuurmans}.} \bibinfo{year}{2020}\natexlab{}.
\newblock \showarticletitle{GenDICE: Generalized Offline Estimation of Stationary Values}. In \bibinfo{booktitle}{\emph{International Conference on Learning Representations}}.
\newblock
\urldef\tempurl%
\url{https://openreview.net/forum?id=HkxlcnVFwB}
\showURL{%
\tempurl}


\end{thebibliography}

\newpage
\appendix
\onecolumn

\section{Hyperparameters}\label{sec:appendix}

The hyperparameters for the experiments on the performance evaluation are shown in \Cref{tab:offline_hyperparameters}.
For D-MOMD, we provide hyperparameters in \Cref{tab:online_hyperparameters}.
\begin{table}[ht]
    \centering
    \begin{tabular}{|l|c|l|}
        \hline
        \textbf{Hyperparameter}        & \textbf{Value} & \textbf{Description} \\ \hline
        Learning Rate                  & 0.001          & Step size for gradient updates  \\ \hline
        $\tau$                    & 20.0           & Temperature \\ \hline
        $\alpha$                          & 0.99           & Weight for momentum in updates \\ \hline
        $\gamma$                & 0.99           & Discount factor \\ \hline
        $\eta$                  & 3.0            & Regularization term for CQL \\ \hline
        B              & 2000           & Number of batches \\ \hline
        N                     & 512            & Batch size \\ \hline
    \end{tabular}
    \caption{Hyperparameters for \algname.}
    \label{tab:offline_hyperparameters}
\end{table}

\begin{table}[ht]
    \centering
    \begin{tabular}{|l|c|l|}
        \hline
        \textbf{Hyperparameter}        & \textbf{Value}  & \textbf{Description} \\ \hline
        Buffer Size                    & 100,000         & Size of the replay buffer\\ \hline
        N            & 256             & Number of samples drawn from the buffer per update \\ \hline
        Update steps             & 4000            & Gradient updates per iteration \\ \hline
        Learning Rate                  & 0.001           & Step size for gradient updates \\ \hline
        $\epsilon$ (start)                & 1.0             & Initial value of $\epsilon$-greedy exploration \\ \hline
        $\epsilon$ (finish)                 & 0.1             & Final value of $\epsilon$ \\ \hline
        $\epsilon$ anneal time            & 1,000,000       & Number of steps to anneal epsilon from start to finish \\ \hline
        $\tau$                    & 20.0            &Temperature \\ \hline
        $\alpha$                          & 0.99            & Weight for momentum in updates \\ \hline
        $\gamma$               & 0.99            & Discount Factor  \\ \hline
        Target Update Interval         & 200             & Steps between target network updates \\ \hline
        Training Interval              & 10              & Number of environment steps per update \\ \hline
    \end{tabular}
    \caption{Hyperparameters for D-MOMD}
    \label{tab:online_hyperparameters}
\end{table}

\end{document}